%% file: main.tex
\documentclass{article}
\usepackage{main,times}
\iclrfinalcopy
\input{math_commands.tex}

\usepackage{hyperref}
\usepackage{url}
\usepackage{xcolor}
\usepackage{booktabs}
\usepackage{pifont}
\usepackage{graphicx}
\usepackage{tabularx}
\usepackage{array}
\usepackage{subcaption}

\title{LIVIN: Benchmarking Spatial and Embodied Intelligence in Digital Twins of Lived-In Homes}

\author{
\hspace*{-0.2em}Peijun Xu\textsuperscript{1,}\thanks{Equal contribution.} \quad
Chuansen Nie\textsuperscript{1,}\footnotemark[1] \quad
Yiyang He\textsuperscript{1,}\footnotemark[1] \quad
Yinuo Bai\textsuperscript{1,2} \quad
Jingyang Liu\textsuperscript{1} \quad
Kuixiang Shao\textsuperscript{1} \\
Yuyang Jiao\textsuperscript{1} \quad
Kuanhao Xia\textsuperscript{1} \quad
Jiayi Zhu\textsuperscript{1} \quad
Zitian Yang\textsuperscript{1} \quad
Yanqi Zhang\textsuperscript{1} \quad
Tianye Tan\textsuperscript{1} \\
Shuwei Di\textsuperscript{1} \quad
Junyi Xu\textsuperscript{1} \quad
Jingyi Yu\textsuperscript{1} \quad
Jiayuan Gu\textsuperscript{1,}\thanks{Corresponding author.}
\\[3mm]
\textsuperscript{1} ShanghaiTech University \quad
\textsuperscript{2} Deemos Technology
\\[2mm]
{\small\texttt{\{xupj2025,niechs2025,heyy2025\}@shanghaitech.edu.cn}}
}

\newcommand{\cmark}{\ding{51}}
\newcommand{\xmark}{\ding{55}}
\begin{document}
\maketitle
\begin{abstract}

Realistic household simulation must capture not only diverse environments but also the lived-in object arrangements and spatial constraints that shape robot motion and interaction. Existing resources often trade off scale, real-world correspondence, and interaction readiness, leaving a gap in faithful, interactive replicas of how real homes are actually arranged. To this end, we introduce \textsc{LIVIN}, a benchmark for spatial and embodied intelligence built on digital twins of 30 diverse lived-in homes. These replicas preserve observed room layouts, furniture configurations, and everyday belongings. To construct them, we design a human-in-the-loop workflow comprising instance recognition, architectural reconstruction, and object generation and placement, with intermediate results reviewed and corrected by humans against the source observations at each stage. We evaluate four tasks in \textsc{LIVIN}: 3D detection, 3D reconstruction, navigation, and loco-manipulation. Our evaluations show that current methods remain challenged by the dense object arrangements, occlusions, limited free space, and constrained interaction regions found in realistic lived-in homes. We hope \textsc{LIVIN} will help advance embodied AI in real-world homes, from spatial understanding to robotic interaction, and ultimately bring embodied intelligence into everyday home environments.

\end{abstract}

\begin{figure}[ht] 
    \centering
    \includegraphics[width=\linewidth]{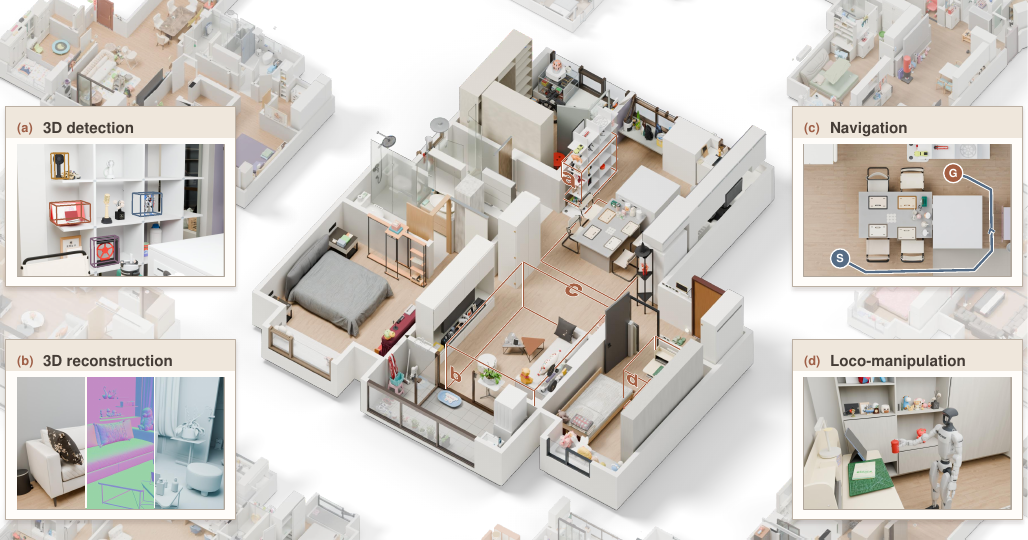}
    \caption{\textbf{Overview of LIVIN.} Built from 30 lived-in homes, \textsc{LIVIN} captures realistic household layouts, furnishings, and daily-use objects, enabling evaluation of 3D detection, reconstruction, navigation, and loco-manipulation in realistic domestic environments.}
    \label{fig:teaser}
\end{figure}

\input{sections/introduction_1}
\input{sections/related-work_2}
\input{sections/dataset_3}

\input{sections/construction_4}
\input{sections/benchmark_3d_5}

\input{sections/benchmark_embodied_6}
\input{sections/conclusion_7}

\bibliography{reference}
\bibliographystyle{reference}

\newpage
\appendix
\input{sections/appendix_dataset}

\input{sections/appendix_construction}
\input{sections/appendix_review}

\input{sections/appendix_reconstruction}
\input{sections/appendix_navigation}
\input{sections/appendix_loco-manipulation}

\end{document}

%% file: math_commands.tex
\usepackage{amsmath,amsfonts,bm}

\def\eqref#1{equation~\ref{#1}}

\def\1{\bm{1}}

\DeclareMathAlphabet{\mathsfit}{\encodingdefault}{\sfdefault}{m}{sl}
\SetMathAlphabet{\mathsfit}{bold}{\encodingdefault}{\sfdefault}{bx}{n}



%% file: sections/introduction_1.tex
\section{Introduction}
\label{sec:introduction}

Everyday homes impose dense and highly variable spatial constraints on robot behavior: furniture limits traversable space, belongings create clutter and occlusion, and surrounding objects affect how a target can be reached or manipulated. Household robots therefore require both \emph{spatial intelligence} to understand objects, geometry, and spatial relationships, as well as \emph{embodied intelligence} to act within these constraints. Environmental diversity is crucial for generalization; for example, $\pi_{0.5}$ reports improved performance as mobile-manipulation training scales from 3 to 104 locations~\citep{pi05}. While simulation can scale such experience, constructing realistic, interactive replicas of diverse lived-in homes remains challenging.

Existing resources address different aspects of the problem. Artist-designed datasets and procedural generators provide diverse layouts and object configurations~\citep{front3d,procthor,infinigenindoors}, but are synthesized rather than grounded in observed homes. Other efforts such as MetaScenes and InternScenes~\citep{metascenes,internscenes} convert real-world scans into simulatable scenes with detailed object arrangements. These pipelines, however, rely heavily on asset retrieval and provide limited coverage of articulated household objects. More recently, agentic systems such as SceneSmith~\citep{scenesmith} automate increasing portions of scene and asset construction. Together, these advances make it possible to move beyond approximate scene replicas toward observation-grounded digital twins that preserve \emph{real household layouts, rich object instances, and interactable articulation}, enabling spatial understanding and embodied behavior to be evaluated in the challenging realistic environments.

We introduce \textbf{LIVIN}, a benchmark built on digital twins of \textbf{30 diverse lived-in homes} for jointly evaluating spatial and embodied intelligence. LIVIN contains 319 rooms and 12,808 object instances, including 1,559 articulated objects, reflecting the dense and varied object arrangements of real households (Sec.~\ref{sec:dataset}). Each home is reconstructed from floorplans, posed multi-view images, and metric scans, preserving room-scale geometry, furniture layouts, and everyday belongings, including small objects that create realistic occlusions and constrain navigation and interaction.

To construct these observation-grounded replicas, we develop a human-in-the-loop workflow comprising instance recognition, architectural reconstruction, and object generation and placement. We treat articulated and non-articulated objects with different reconstruction strategies. Articulated objects are jointly modeled and placed by GPT-6 Astra~\citep{gpt6} through Blender code, preserving their kinematic structure while fitting the surrounding environment. Non-articulated objects are generated by Rodin~\citep{clay} from enhanced reference images and then placed to match the captured arrangements. Each construction stage is followed by human review, allowing intermediate results to be inspected and corrected against the source observations before proceeding.

\textbf{LIVIN} supports four benchmarks: 3D detection, 3D reconstruction, navigation, and loco-manipulation. The 3D benchmarks evaluate object localization and geometry recovery under clutter, occlusion, and dense household arrangements. The navigation benchmark targets long-horizon humanoid navigation, requiring robots to reason about routes and execute physically feasible whole-body motion through constrained spaces. The loco-manipulation benchmark further evaluates physical interaction, including pick-and-place, pushing, articulated-object manipulation, and long-horizon tasks that couple locomotion with dependent manipulation subgoals. Across these benchmarks, we find that existing methods still struggle in complex household scenes: 3D methods may miss objects or recover inaccurate geometry, while embodied systems may fail to localize the intended destination, escape obstacles, establish stable grasps, or complete long-horizon manipulation sequences.

The work is organized around three contributions:
\begin{itemize}
    \item \textbf{LIVIN-Scenes: 30 diverse lived-in homes}, providing observation-grounded digital twins that preserve realistic household layouts, everyday object arrangements, and rich object categories under natural spatial constraints.
    
    \item \textbf{A human-in-the-loop reconstruction workflow grounded in captured observations}, combining agentic reasoning, code-based modeling, and image-conditioned 3D generation to reconstruct architectural structure, articulated objects, and everyday belongings with stage-wise human verification.
    
    \item \textbf{LIVIN-Bench: a unified evaluation suite for 3D and embodied tasks}, spanning 3D detection, 3D reconstruction, navigation, and loco-manipulation to evaluate scene understanding, robot mobility, and physical interaction in realistic household environments.
\end{itemize}

%% file: sections/related-work_2.tex
\section{Related Work}
\label{sec:related-work}

\paragraph{Real-to-Sim Generation.}
Recent 3D generative models have substantially improved the conversion of real-world observations into reusable virtual assets and structured scenes. At the object level, methods such as TRELLIS, Clay, and SAR3D recover detailed geometry and appearance from sparse visual inputs~\citep{xiang2025structured,xiang2026native,clay,chen2025sar3d}, while some works further extend 3D generation to articulated and interaction-ready assets with explicit part structures and motion parameters~\citep{liu2024cage,zhou2026articraft,gao2025meshart,wang2026artllm}. At the scene level, recent methods such as MIDI and SceneGen generate compositional 3D environments by jointly modeling object geometry and scene layout from visual or structured conditions~\citep{huang2025midi,meng2026scenegen}, while CAST, SAM3D and RecGen reconstruct 3D scenes by recovering object geometry and estimating object poses from visual observations~\citep{yao2025cast,sam3d,zadaianchuk2026reconstruction}. \textsc{LIVIN} extends this direction to complete lived-in homes through a human-in-the-loop agentic workflow that reconstructs architectural structure, scene layout, articulated objects, and high-fidelity everyday objects, producing more complete and faithful replicas of real household environments.

\paragraph{Indoor Scene Datasets.}
Existing indoor scene resources are built through several distinct approaches. Artist-designed datasets such as 3D-FRONT and HSSD offer high-quality structured interiors but require substantial manual effort~\citep{front3d,hssd}. Procedural methods such as ProcTHOR and Infinigen Indoors scale more easily~\citep{procthor,infinigenindoors}, but their rule-based layouts and reused assets limit correspondence to real homes. Recent agentic and generative systems further automate scene creation~\citep{holodeck,xia2026sage,scenesmith,lucida,lumera}, but their outputs still require verification for scene-level consistency. Capture-grounded datasets such as Scan2CAD, ReplicaCAD, MetaScenes, and InternScenes derive virtual environments from real-world observations~\citep{scan2cad,habitat2,metascenes,internscenes}, yet retrieval, rearrangement, or partial reconstruction may not fully preserve instance-specific structure and complete object arrangements. In contrast, \textsc{LIVIN} is fully grounded in real lived-in homes, with reconstruction initialized by agents and subsequently aligned and corrected against real-world evidence, balancing efficiency and fidelity to produce detailed, articulated, interaction-ready replicas.

\paragraph{Navigation and Loco-manipulation Benchmarks.}
Existing embodied benchmarks span real scans, curated scenes, and designed or procedurally generated environments. VLN-CE and Interactive Gibson build on scanned homes, with the latter adding movable objects for interactive navigation~\citep{vlnce,interactivegibson}. GRUtopia and HumanoidVLN evaluate navigation in large-scale constructed or reconstructed environments~\citep{grutopia,humanoidvln}, while BEHAVIOR-1K, RoboCasa365, HomeRobot, BiGym, HumanoidBench, and SIMPLE cover manipulation and whole-body control across curated or generated scenes~\citep{behavior1k,robocasa365,yenamandra2023homerobot,bigym,humanoidbench,wei2026simple}. In contrast, \textsc{LIVIN} preserves the observed layouts, furniture, everyday belongings, and articulated fixtures of individual lived-in homes, retaining naturally occurring spatial constraints that jointly affect navigation and manipulation.

%% file: sections/dataset_3.tex
\section{\textsc{LIVIN} Benchmark}
\label{sec:dataset}

\textsc{LIVIN} contains 30 digital twins of lived-in homes, comprising 319 rooms and 12808 specific object instances, including 1559 articulated objects. The collection covers a broad range of household scales, room functions, furnishing styles, and object compositions. We assess scenes in \textsc{LIVIN} along two dimensions: fidelity to the captured homes and preservation of real-world household complexity. Additional dataset statistics and qualitative examples are provided in Appendix~\ref{app:dataset-statistics}.

\paragraph{Real-world correspondence.}
Each \textsc{LIVIN} scene is reconstructed from a specific lived-in home and grounded in its captured evidence. At the scene level, \textsc{LIVIN} preserves the architectural layout and fine-grained details of each home, including ceilings, niches, beams, and wall openings. At the object level, it provides instance-specific meshes and explicitly models the movable parts and kinematic relationships of articulated household objects. We compare \textsc{LIVIN} with other indoor scene datasets in Tab.~\ref{tab:scene_comparison} in terms of scene- and object-level properties. We further validate scene quality through a user study on physical plausibility, perceived realism, and reconstructed mesh quality.  \textsc{LIVIN} receives consistently higher ratings across all three dimensions than the compared datasets, reflecting its high-fidelity reproduction of real-home structure, object arrangements details.

\paragraph{Reproducing household complexity.}
We further quantify how well \textsc{LIVIN} captures the object-level complexity of real household environments in Tab.~\ref{tab:scene_comparison}. The results show that \textsc{LIVIN} achieves the highest object count and density among datasets with available statistics, while also containing a larger proportion of articulated objects than datasets reporting this metric. Together, these findings suggest that \textsc{LIVIN} better captures the object richness and interaction-relevant structure of lived-in household environments.

\begin{table}[t]
    \centering
    \caption{
    \textbf{Comparison with existing indoor scene datasets.}
    Inst.-Spec. Mesh denotes instance-specific object meshes, and Artic. Struct. denotes modeled articulation. Obj. Dens. denotes object density (obj./m$^2$), and Artic. Obj. denotes the percentage of articulated objects. Plaus., Realism, and Mesh Qual. denote human preference rates for physical plausibility, scene realism, and mesh quality.
    }
    \label{tab:scene_comparison}
    \scriptsize
    \setlength{\tabcolsep}{1.5pt}
    \renewcommand{\arraystretch}{1.12}

    \begin{tabular*}{\linewidth}{
        @{\extracolsep{\fill}}
        lccccccccc
        @{}
    }
        \toprule
        &
        \multicolumn{3}{c}{\textbf{Properties}}
        &
        \multicolumn{3}{c}{\textbf{Statistics}}
        &
        \multicolumn{3}{c}{\textbf{User Study}}
        \\
        \cmidrule(lr){2-4}
        \cmidrule(lr){5-7}
        \cmidrule(lr){8-10}

        \textbf{Dataset}
        & \textbf{Real-World}
        & \textbf{Inst.-Spec.}
        & \textbf{Artic.}
        & \textbf{Avg. Obj.}
        & \textbf{Obj. Dens.}
        & \textbf{Artic. Obj.}
        & \textbf{Plaus.} $\uparrow$
        & \textbf{Realism} $\uparrow$
        & \textbf{Mesh Qual.} $\uparrow$
        \\

        &
        \textbf{Grounding}
        & \textbf{Mesh}
        & \textbf{Struct.}
        & \textbf{/ Room}
        & \textbf{(obj./m$^2$)}
        & \textbf{(\%)}
        & \textbf{(\%)}
        & \textbf{(\%)}
        & \textbf{(\%)}
        \\
        \midrule

        3D-FRONT~\citep{front3d}
        & \xmark & \xmark & \xmark
        & 6.93 & 0.39 & --
        & 15.33 & 10.10 & 12.60
        \\

        ProcTHOR~\citep{procthor}
        & \xmark & \xmark & \cmark
        & 18.33 & 0.82 & 11.11
        & 5.23 & 8.36 & 2.03
        \\

        HSSD~\citep{hssd}
        & \xmark & \xmark & \cmark
        & 20.17 & 1.58 & 10.08
        & 13.24 & 21.26 & 18.70
        \\

        \midrule

        ReplicaCAD~\citep{habitat2}
        & \cmark & \xmark & \cmark
        & 30.48 & 0.38 & 5.38
        & 10.80 & 4.18 & 3.66
        \\

        MetaScenes~\citep{metascenes}
        & \cmark & \xmark & \xmark
        & 21.76 & 1.63 & --
        & 4.53 & 1.05 & 2.03
        \\

        \midrule

        \textbf{\textsc{LIVIN}}
        & \cmark & \cmark & \cmark
        & \textbf{40.15}
        & \textbf{3.70}
        & \textbf{12.17}
        & \textbf{50.87}
        & \textbf{55.05}
        & \textbf{60.98}
        \\

        \bottomrule
    \end{tabular*}
\end{table}

%% file: sections/construction_4.tex
\section{Replica Construction with Human Verification}
\label{sec:construction}

Given the floorplan, posed multi-view images, and a metric scan of a real home, we develop a human-in-the-loop agentic pipeline for constructing its digital replica. We use GPT-6 Astra~\citep{gpt6} as the reconstruction agent throughout the pipeline. The workflow consists of three stages: instance recognition, architectural reconstruction, and object generation and placement. Each stage is followed by human inspection and refinement to ensure consistency with the captured environment. Additional implementation details are provided in Appendix~\ref{app:automation}.

\subsection{Instance Recognition}
\label{sec:asset-recognition}

We first establish a complete inventory of the objects present in each home to identify the instances to be reconstructed. The agent identifies and segments object instances across posed multi-view images, revisiting densely populated regions and consulting complementary views to recover objects that are partially or fully occluded in individual images. Cross-view detections are matched against the existing asset inventory to remove duplicates. For each instance, we record its semantic label, material attributes, associated room, and articulation properties. The articulation information determines how the object is generated in the next stage. The agent also selects a reference view of each object for subsequent generation, based on visibility, completeness, and image quality.

\subsection{architectural Construction}
\label{sec:structure-creation}

Reconstruction begins with the architectural structure of each home. Based on the 2D floorplan, metric scans, and available room metadata, we first build a coarse 3D architectural shell with the major walls, floors, ceilings, and openings. More details of the initial shell construction are provided in Appendix~\ref{app:agent-protocol}. The agent then aligns this initial structure with the posed images, refines wall connections and openings, and adds observed architectural details such as niches, beams, and ceiling features. Windows are further reconstructed with frames and glass according to their observed configurations.

\subsection{Object Generation and Placement}
\label{sec:asset-creation}

For object reconstruction and placement, we follow the generation routes determined during instance recognition and use different strategies for articulated and non-articulated objects.

\paragraph{Articulated objects.}
For articulated objects, the agent directly generates Blender code for in-context construction. It estimates each object's pose from the selected reference view and surrounding room geometry, while jointly modeling its geometry, materials, articulation, and placement. Coupling generation with placement allows the reconstructed geometry to conform to the observed footprint and surrounding structures. Functional components such as storage cavities and drawer boxes are also preserved, with movable parts explicitly represented through motion axes and kinematic relationships.

\paragraph{Non-articulated objects.}
For non-articulated objects, we use Rodin~\citep{clay} to generate the corresponding 3D assets. Given the selected reference view of each object, we use GPT-Image-2~\citep{gptimage2} to generate a clean product-style image by removing occlusions and unrelated surroundings while preserving the object's appearance and constituent parts. Rodin then generates a high-fidelity mesh from this cleaned reference. The agent subsequently estimates each object's scale, pose, and placement from the constructed scene context and source observations. The placement process is required to preserve spatial relations such as stacking, containment, wall mounting, and suspension, while avoiding inter-object collisions to improve physical plausibility.

\subsection{Human Review and Refinement}
\label{sec:human-review}

Automatic reconstruction may introduce errors that require human correction. We therefore review and refine the output of each construction stage before proceeding to the next, limiting error propagation. During instance recognition, reviewers correct detection errors, room assignments, and reference images for subsequent Rodin generation. For scene reconstruction, we develop an interactive review interface that enables reviewers to compare the reconstructed scene against the source observations and correct errors in geometry, articulation, appearance, and placement. For architectural elements and articulated objects, reviewers select the relevant structure or part and provide natural-language instructions to precisely guide the reconstruction agent toward the required corrections. Defective non-articulated objects can instead be regenerated with Rodin, while object poses and placements can be adjusted directly through the interface. Further details of the review and correction process are provided in Appendix~\ref{app:annotation}.

%% file: sections/benchmark_3d_5.tex
\section{3D Detection and Reconstruction}
\label{sec:benchmark-3d}

\begin{figure}[htb]
    \centering
    \includegraphics[width=\linewidth]{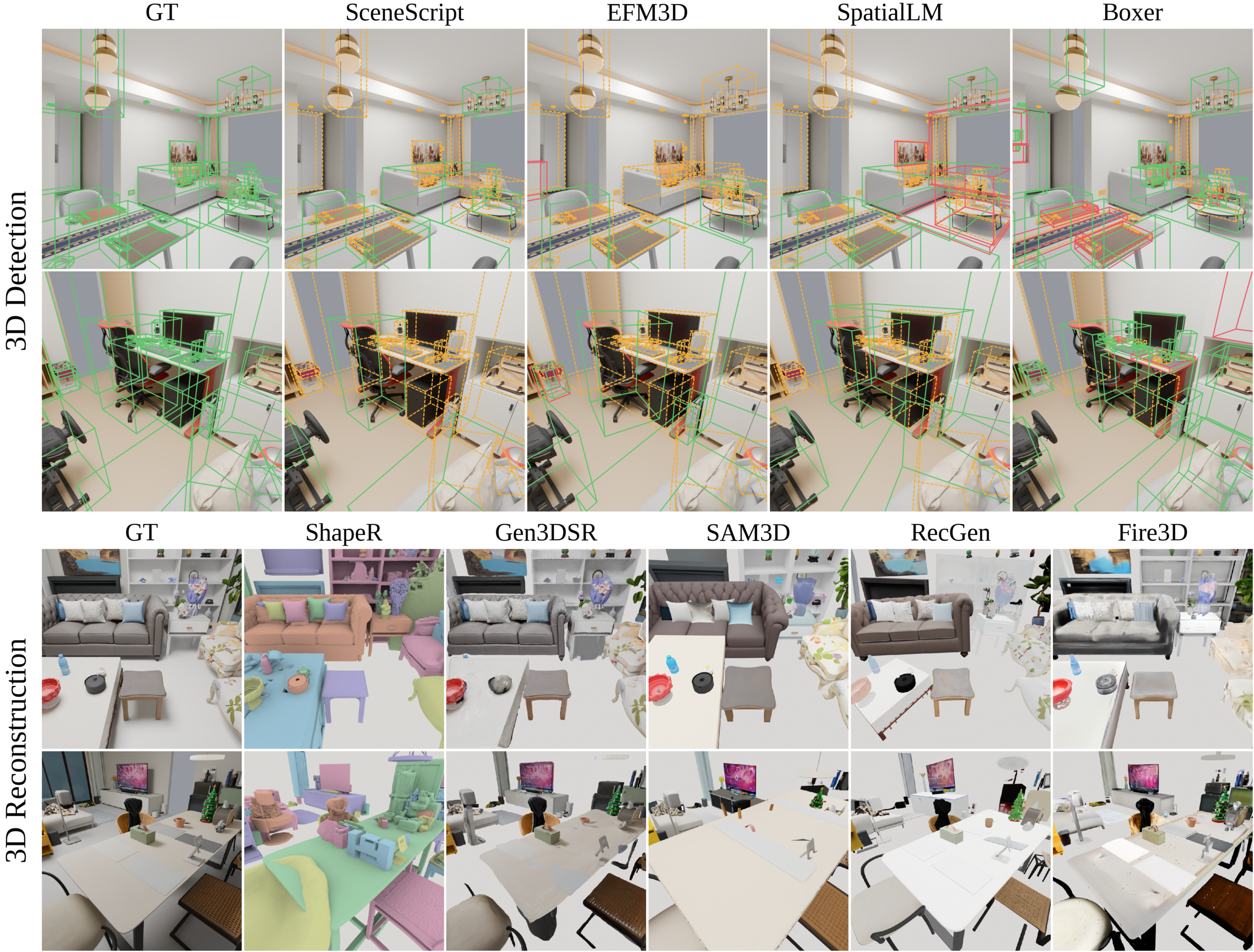}

    \caption{\textbf{Qualitative results for 3D detection and reconstruction.}
Top: 3D detection predictions from SceneScript, EFM3D, SpatialLM, and Boxer compared with ground truth, where green boxes denote correct detections, yellow boxes indicate missed objects, and red boxes denote incorrect predictions.
Bottom: 3D reconstructions from ShapeR, Gen3DSR, SAM3D, RecGen, and Fire3D compared with ground-truth.}
    \label{fig:benchmark-3d}
    
\end{figure}

We consider two 3D tasks in the captured household environments: object detection and scene reconstruction. These tasks evaluate object recognition, localization, and geometry recovery under occlusion and dense object arrangements. Fig.~\ref{fig:benchmark-3d} presents representative qualitative results for both tasks, comparing model predictions and reconstructions with the corresponding ground truth.

\begin{table}[t]
    \centering
    \caption{
    \textbf{Quantitative evaluation of 3D detection and reconstruction on \textsc{LIVIN}.}
    F-Score-S uses a surface-distance threshold of $0.05\,\mathrm{m}$, and F1 uses a 3D box-matching threshold of $0.25$.
    }
    \label{tab:3d_tasks}
    \scriptsize
    \setlength{\tabcolsep}{1.5pt}
    \renewcommand{\arraystretch}{1.12}

    \begin{tabular*}{\linewidth}{
        @{\extracolsep{\fill}}
        lccccc
        @{}
    }
        \toprule
        \multicolumn{6}{c}{\textbf{3D Detection}} \\
        \midrule

        \textbf{Method}
        & \textbf{mAP} $\uparrow$
        & \textbf{Prec.@0.25/0.5} $\uparrow$
        & \textbf{Recall@0.25/0.5} $\uparrow$
        & \textbf{F1@0.25/0.5} $\uparrow$
        & \textbf{3D IoU@0.25/0.5} $\uparrow$
        \\
        \midrule

        SceneScript~\citep{avetisyan2024scenescript}
        & --
        & 0.608 / 0.428
        & 0.083 / 0.058
        & 0.146 / 0.103
        & 0.591 / 0.678
        \\

        EFM3D~\citep{straub2024efm3d}
        & 0.071
        & 0.617 / 0.362
        & 0.096 / 0.056
        & 0.166 / 0.097
        & 0.541 / 0.657
        \\

        SpatialLM~\citep{SpatialLM}
        & --
        & \textbf{0.743 / 0.522}
        & 0.111 / 0.078
        & 0.192 / 0.135
        & \textbf{0.601 / 0.698}
        \\

        Boxer~\citep{boxer2026}
        & \textbf{0.306}
        & 0.721 / 0.477
        & \textbf{0.417 / 0.276}
        & \textbf{0.528 / 0.349}
        & 0.568 / 0.659
        \\

        \bottomrule
    \end{tabular*}

    \vspace{2mm}

    \begin{tabular*}{\linewidth}{
        @{\extracolsep{\fill}}
        lcccccc
        @{}
    }
        \toprule
        \multicolumn{7}{c}{\textbf{3D Reconstruction}} \\
        \midrule

        \textbf{Method}
        & \textbf{CD-S} $\downarrow$
        & \textbf{F-Score-S@0.05m} $\uparrow$
        & \textbf{F1@0.25} $\uparrow$
        & \textbf{PSNR} $\uparrow$
        & \textbf{SSIM} $\uparrow$
        & \textbf{LPIPS} $\downarrow$
        \\
        \midrule

        Gen3DSR~\citep{Ardelean2025Gen3DSR}
        & 0.214
        & 0.336
        & 0.627
        & 14.03
        & 0.678
        & 0.369
        \\

        SAM3D~\citep{sam3d}
        & 0.133
        & 0.381
        & 0.685
        & 12.07
        & 0.625
        & 0.460
        \\

        RecGen~\citep{zadaianchuk2026reconstruction}
        & 0.185
        & 0.431
        & 0.664
        & 12.06
        & 0.647
        & 0.428
        \\
        
        ShapeR~\citep{Siddiqui_2026_CVPR}
        & 0.040
        & 0.602
        & 0.838
        & --
        & --
        & --
        \\
        Fire3D~\citep{xia2026fire3d}
        & \textbf{0.022}
        & \textbf{0.732}
        & \textbf{0.880}
        & \textbf{14.88}
        & \textbf{0.701}
        & \textbf{0.360}
        \\

        \bottomrule
    \end{tabular*}
\end{table}

\subsection{3D Object Detection}

We evaluate SceneScript~\citep{avetisyan2024scenescript}, EFM3D~\citep{straub2024efm3d}, SpatialLM~\citep{SpatialLM}, and Boxer~\citep{boxer2026} on 268 rooms from the 30 homes in \textsc{LIVIN} using released checkpoints without fine-tuning. Method-specific inputs are constructed from rendered RGB-D observations: SceneScript and SpatialLM use point clouds, EFM3D uses multi-view RGB with semidense geometry and camera information, and Boxer uses RGB-D frames with OWLv2~\citep{minderer2023scaling} 2D proposals. We evaluate class-agnostic room-level predictions against visible-object oriented boxes using Hungarian matching with 3D IoU, reporting precision, recall, F1, matched-box IoU. For methods that provide confidence scores, we additionally report mAP averaged over IoU thresholds from 0.05 to 0.50 in increments of 0.05. As shown in Tab.~\ref{tab:3d_tasks}, Boxer achieves the highest F1 but still limited recall, while SpatialLM attains the highest precision and matched-box IoU with only 0.111 recall at IoU 0.25. These results show that accurate localization of detected objects does not ensure complete scene coverage, and that object recovery remains a major challenge in \textsc{LIVIN}'s object-dense household scenes.

\subsection{3D Scene Reconstruction}
For the 3D reconstruction task, we evaluate five methods on 128 views sampled from \textsc{LIVIN} scenes, with results reported in Tab.~\ref{tab:3d_tasks}. All methods receive RGB, GT instance masks, depth, and camera calibration. Fire3D~\citep{xia2026fire3d} and ShapeR~\citep{Siddiqui_2026_CVPR} additionally use GT oriented bounding boxes, while ShapeR also takes depth-derived point clouds and object descriptions. ShapeR and Fire3D achieve higher recall and geometric accuracy overall, which may partly reflect the stronger priors provided by their additional GT inputs. SAM3D~\citep{sam3d}, RecGen~\citep{zadaianchuk2026reconstruction}, and Gen3DSR~\citep{Ardelean2025Gen3DSR} show different trade-offs between geometry and appearance. SAM3D achieves lower surface error, while RecGen obtains a higher scene level F-Score. Gen3DSR is less accurate geometrically, with lower scene level F-Score and F1, but better preserves appearance, yielding the strongest PSNR, SSIM, and LPIPS among the three. Additional analysis and discussion are provided in Appendix~\ref{app:3d_reconstruction}.

%% file: sections/benchmark_embodied_6.tex
\begin{figure}[t]
    \centering
    \includegraphics[width=\linewidth]{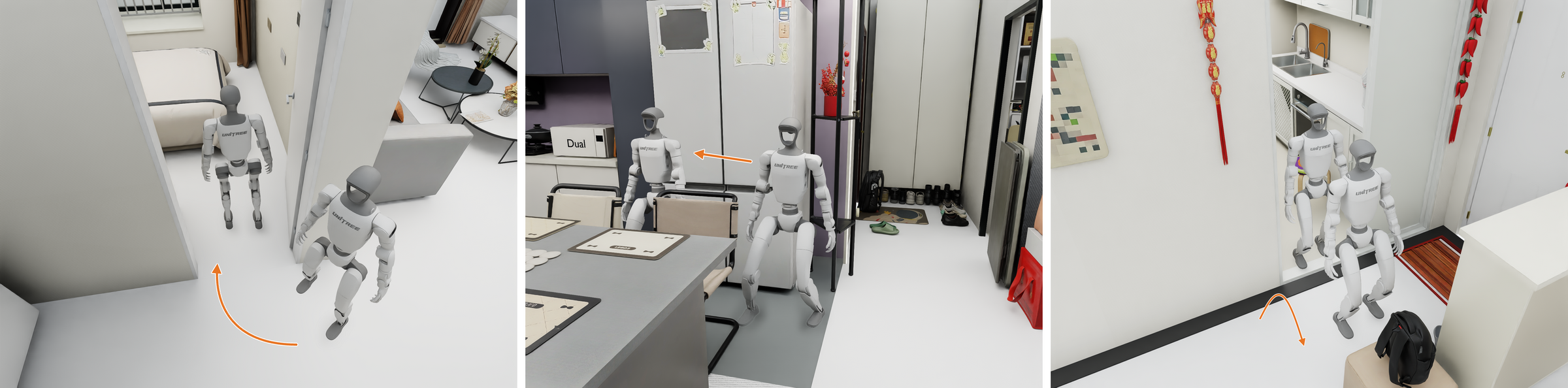}
    \caption{\textbf{Humanoid navigation in \textsc{LIVIN}.}
    \textsc{LIVIN} evaluates humanoid navigation in faithful replicas of real lived-in homes under naturally occurring spatial constraints. From left to right, examples show the robot turning into a room, moving sideways through a narrow passage, and stepping over a doorway threshold.}
    \label{fig:navigation}
\end{figure}

\section{Navigation and Loco-Manipulation}
\label{sec:benchmark}

We evaluate two embodied tasks in \textsc{LIVIN}: navigation and loco-manipulation. Unlike prior benchmarks built on purpose-designed or deliberately cluttered test environments, \textsc{LIVIN} provides faithful replicas of complete lived-in homes, preserving naturally occurring spatial constraints and object arrangements. This enables the evaluation of robot mobility and interaction under conditions that more closely reflect real household operations. Moreover, the diverse structures preserved in \textsc{LIVIN} naturally support a broad range of task settings, including narrow passages, interactive household fixtures and object manipulation across different room types.

\subsection{Navigation}
\label{sec:navigation}

We formulate navigation as a long-horizon vision-language task for physically embodied humanoids, requiring the robot to follow route instructions using visual observations while executing feasible whole-body motion. Fig.~\ref{fig:navigation} visualizes several representative navigation tasks in \textsc{LIVIN}. We compare current general-purpose multimodal models with state-of-the-art VLN methods, evaluating all methods without training or fine-tuning on \textsc{LIVIN}. Specifically, we consider GPT-6 Astra, OmniNav~\citep{xue2026omninav}, and DualVLN~\citep{wei2026ground} as the high-level navigation methods. All methods use GR00T-WBC~\citep{gr00t_wbc} as the controller and the same route instructions, while retaining their native observation interfaces: GPT-6 Astra uses on-demand head-camera RGB with interaction history, OmniNav uses three-view RGB with visual history, and DualVLN uses head-camera RGB with visual history.

Our experiments cover 45 instruction-following routes across 30 \textsc{LIVIN} scenes using a Unitree G1 humanoid, with MuJoCo~\citep{mujoco} for physics simulation and Isaac Sim~\citep{isaacsim} for RGB rendering. Results are reported in Tab.~\ref{tab:navigation-results}. GPT-6 Astra performs best overall, achieving a 41.48\% success rate, the highest SPL, and the lowest navigation error. This may stem from Astra's stronger scene reasoning and information integration: in most trials, the robot maintains the correct motion direction and successfully selects the instructed branch at route intersections. The two specialized VLN baselines exhibit different failure modes. We find that OmniNav sometimes gets stuck in simple collision regions, repeatedly attempting similar motions without escaping, which leads to the highest collision count among the evaluated methods. In contrast, DualVLN recovers more effectively from such collisions but more often drifts from the intended route, resulting in low success and high navigation error. GPT-6 Astra is less affected by these issues, but can still fail to accurately reach the intended region because of imperfect scene understanding in complex household environments, or become limited by local motion planning in tightly constrained spaces. These results expose remaining challenges in current navigation systems under realistic household conditions. Detailed implementation, experimental settings, and further discussion are provided in Appendix~\ref{app:navigation}.

\begin{table}[t]
    \centering
    \caption{
    \textbf{Quantitative comparison of navigation methods on \textsc{LIVIN}.}
    SR denotes success rate and SPL denotes success weighted by path length; both are reported in percent. Collision
      Count is the mean number of robot–environment collision events, and NE denotes the mean geodesic distance from the robot’s final position to the goal in meters.
    }
    \label{tab:navigation-results}
    \small
    \renewcommand{\arraystretch}{1.12}

    \begin{tabular*}{\linewidth}{
        @{\extracolsep{\fill}}
        lcccc
        @{}
    }
        \toprule
        \textbf{Method}
        & \textbf{SR (\%)} $\uparrow$
        & \textbf{SPL (\%)} $\uparrow$
        & \textbf{Collision Count} $\downarrow$
        & \textbf{NE (m)} $\downarrow$
        \\
        \midrule

        OmniNav~\citep{xue2026omninav}
        & 6.67
        & 3.66
        & 80.59
        & 6.69
        \\

        DualVLN~\citep{wei2026ground}
        & 4.44
        & 2.56
        & \textbf{42.11}
        & 5.97
        \\

        GPT-6 Astra~\citep{gpt6}
        & \textbf{41.48}
        & \textbf{23.59}
        & 55.41
        & \textbf{2.16}
        \\

        \bottomrule
    \end{tabular*}
\end{table}

\subsection{Loco-Manipulation}
\label{sec:locomanipulation}
\begin{figure}[htbp]
    \centering
    \includegraphics[width=\linewidth]{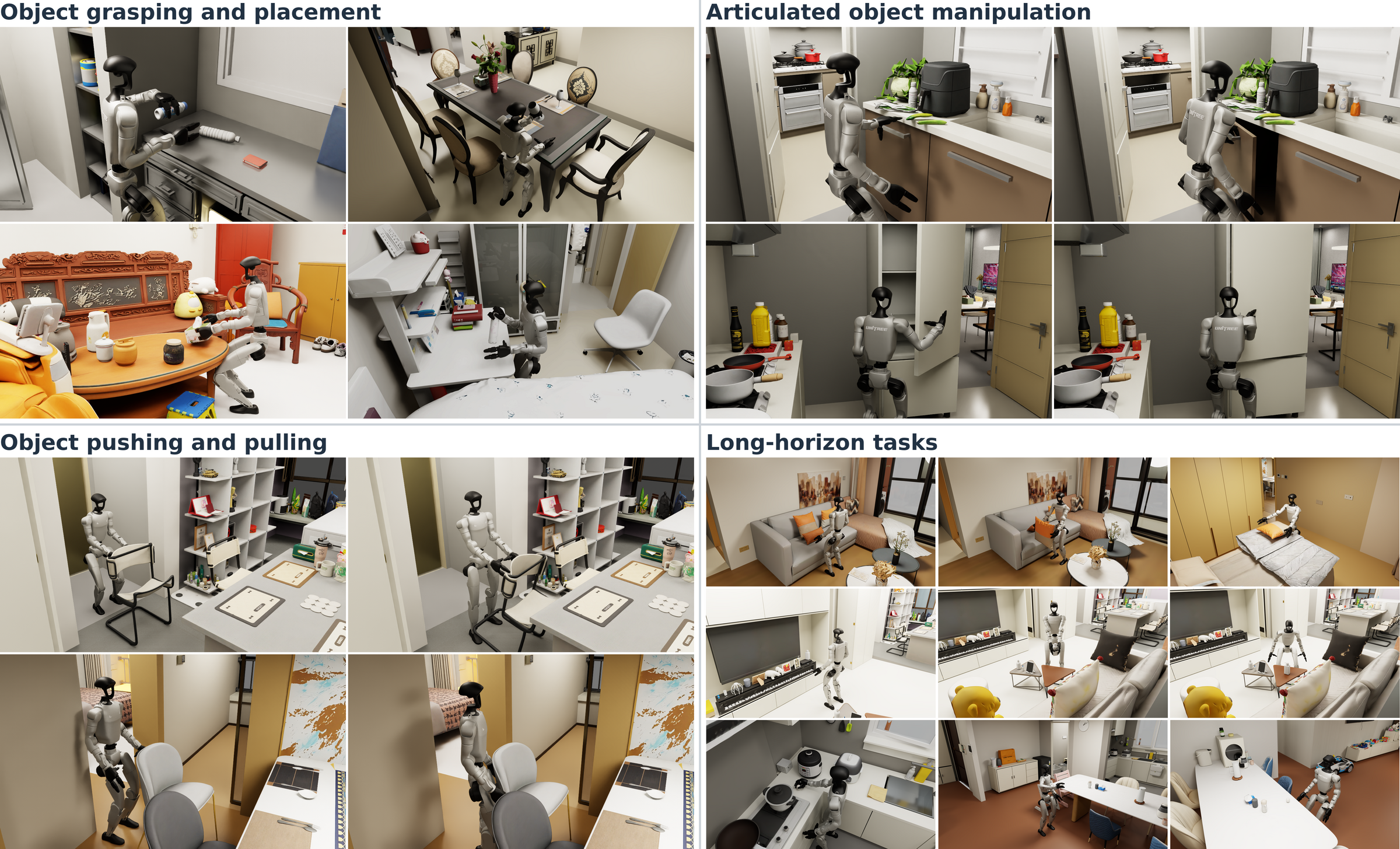}
    \caption{\textbf{Loco-manipulation tasks.}
    Examples of object grasping and placement, object pushing and pulling,
    articulated-object manipulation, and long-horizon tasks in household environments.}
    \label{fig:locomanipulation-tasks}
\end{figure}

We evaluate instruction-conditioned household loco-manipulation, requiring a humanoid to coordinate locomotion and object interaction under the spatial constraints of \textsc{LIVIN}'s lived-in homes. Tasks span four families: \textbf{object grasping and placement}, \textbf{object pushing and pulling}, \textbf{articulated-object manipulation}, and \textbf{long-horizon tasks} with dependent subgoals. Examples include grasping and placing objects in constrained spaces, moving obstructions to access targets or clear passages, and opening a cabinet before placing an object inside (Fig.~\ref{fig:locomanipulation-tasks}). Household arrangements therefore form part of the task rather than merely its background: robots must operate within existing spatial constraints and, when necessary, modify the environment to enable subsequent locomotion or manipulation. We compare two multimodal large language models (MLLMs), DeepSeek-V4.1-Flash~\citep{deepseekv41flash} and GPT-6 Astra~\citep{gpt6}, with the humanoid foundation model $\Psi_0$~\citep{wei2026psi_0}, evaluating all methods without training or fine-tuning on \textsc{LIVIN} (Tab.~\ref{tab:locomanipulation-results}).

We evaluate 35 tasks across 30 \textsc{LIVIN} homes using a Unitree G1 humanoid with Dex3 hands and the physics and rendering backends described in Sec.~\ref{sec:navigation}. Both MLLMs receive task instructions, head-camera RGB images, and robot state information, and control the robot through a shared SONIC-based whole-body controller~\citep{luo2026sonic}. $\Psi_0$ uses its released task experts and native controllers. Each method is evaluated in three trials per task from the same initial configuration. We report success rate (SR) across all task families and mean process score (PS) for long-horizon tasks. Further experimental details are provided in Appendix~\ref{app:loco-protocol}.

\begin{table}[t]
    \centering
    \caption{
    \textbf{Loco-manipulation performance across task families on \textsc{LIVIN}.}
    SR denotes end-to-end success rate in percent and is reported for all four task families.
    PS denotes the mean process score across trials (0--100) and is reported only for long-horizon tasks.
    Higher is better.
    }
    \label{tab:locomanipulation-results}
    \fontsize{8}{10}\selectfont
    \setlength{\tabcolsep}{2pt}
    \renewcommand{\arraystretch}{1.12}

    \begin{tabular*}{\linewidth}{
        @{\extracolsep{\fill}}
        lccccc
        @{}
    }
        \toprule
        \textbf{Method}
        & \shortstack{\textbf{Grasping}\\\textbf{and placement}}
        & \shortstack{\textbf{Pushing}\\\textbf{and pulling}}
        & \shortstack{\textbf{Articulated-object}\\\textbf{manipulation}}
        & \multicolumn{2}{c}{\shortstack{\textbf{Long-horizon tasks}}}
        \\
        \cmidrule(lr){5-6}

        & \textbf{SR (\%)} $\uparrow$
        & \textbf{SR (\%)} $\uparrow$
        & \textbf{SR (\%)} $\uparrow$
        & \textbf{SR (\%)} $\uparrow$
        & \textbf{PS} $\uparrow$
        \\
        \midrule

        DeepSeek-V4.1-Flash~\citep{deepseekv41flash}
        & 6.7
        & 20.0
        & 19.0
        & 0.0
        & 2.8
        \\

        GPT-6 Astra~\citep{gpt6}
        & \textbf{24.4}
        & \textbf{33.3}
        & \textbf{61.9}
        & \textbf{12.5}
        & \textbf{15.9}
        \\

        $\Psi_0$~\citep{wei2026psi_0}
        & 0.0
        & 0.0
        & 0.0
        & 0.0
        & 0.0
        \\

        \bottomrule
    \end{tabular*}
\end{table}

\paragraph{Results and analysis.}
As shown in Tab.~\ref{tab:locomanipulation-results}, GPT-6 Astra achieves the highest success rates across all four task families, while DeepSeek-V4.1-Flash achieves limited success and $\Psi_0$ completes no tasks under the evaluated zero-shot configuration. Qualitative observations reveal target-localization errors and unsuccessful recovery attempts for DeepSeek-V4.1-Flash. For GPT-6 Astra, a cup-grasping example illustrates difficulties in transitioning from target approach to manipulation: the robot approaches the target but makes unintended contact during reaching before establishing a stable grasp. Performance also varies across interaction types, with GPT-6 Astra achieving 61.9\% SR on articulated-object manipulation but only 24.4\% on grasping and placement. Long-horizon completion remains limited: GPT-6 Astra achieves 12.5\% SR and a mean PS of 15.9, while DeepSeek obtains a mean PS of 2.8 with no successful completions, indicating limited progress on intermediate physical subgoals. Together, these results expose limitations beyond target approach, in both establishing effective physical contact and completing dependent interactions within household tasks.

%% file: sections/conclusion_7.tex
\section{Conclusion}
\label{sec:conclusion}
We presented \textsc{LIVIN}, a benchmark of digital twins of lived-in homes for spatial and embodied intelligence. By preserving real-world scene structure, object arrangements and articulation, \textsc{LIVIN} enables evaluation across 3D perception, reconstruction, navigation, and loco-manipulation in realistic household environments. Our results show that current methods still struggle with dense object arrangements, frequent occlusion, and physically constrained interactions. By exposing these challenges in realistic household environments, \textsc{LIVIN} provides a testbed for driving progress toward more robust and generalizable embodied systems.

%% file: sections/appendix_dataset.tex
\section{Dataset Statistics}
\label{app:dataset-statistics}

\begin{figure}[!htbp]
    \centering
    \includegraphics[width=\linewidth]{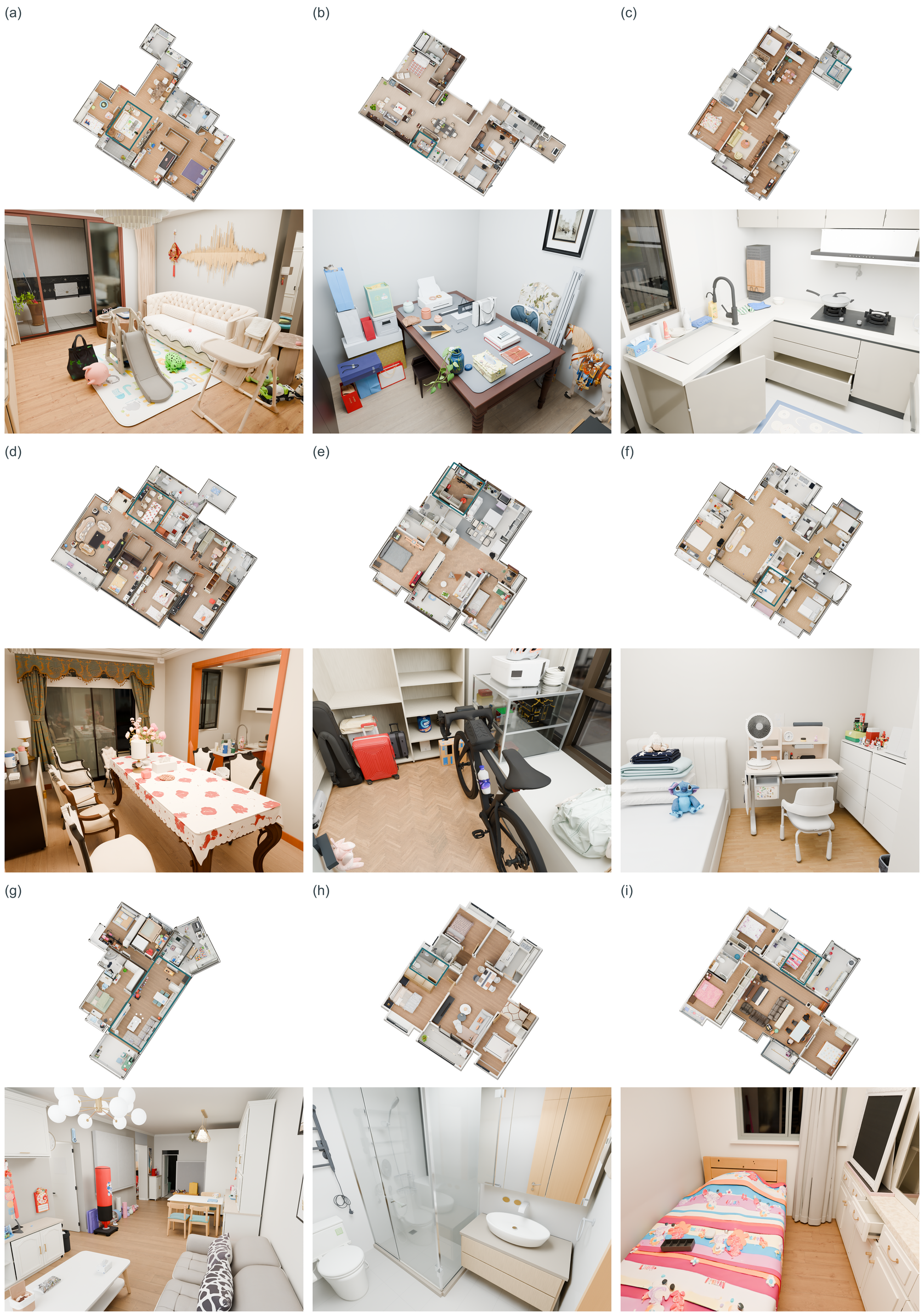}
    \caption{\textbf{Gallery.}
    Top-down view and close-up view.}
    \label{fig:gallery}
\end{figure}

\begin{figure}[ht]
    \centering
    \includegraphics[width=\linewidth]{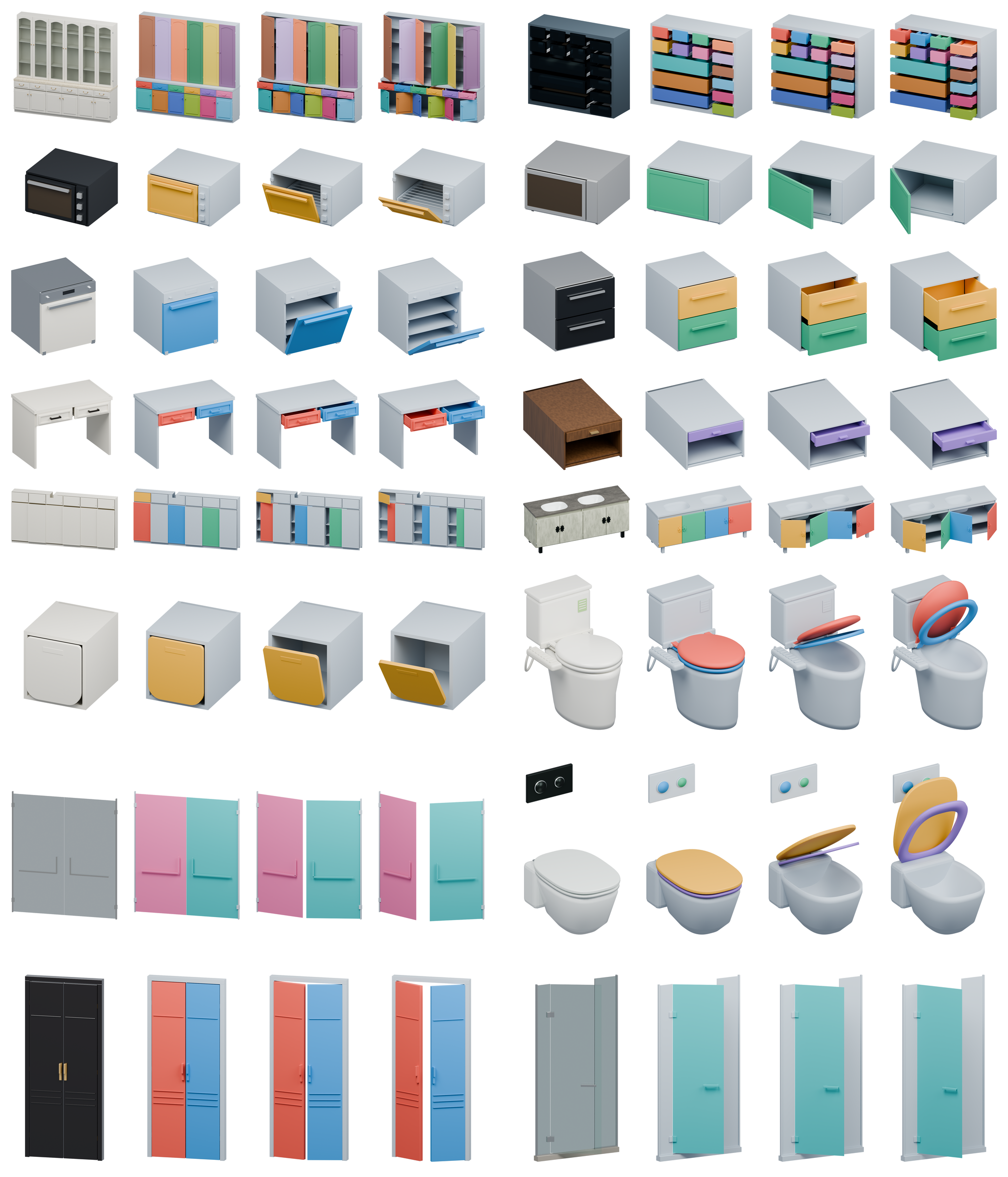}
    \caption{\textbf{Articulated objects in \textsc{LIVIN}}. Each quartet shows original appearance and motion frames.
Colors identify moving assemblies consistently within each object; gray denotes stationary structure.}
    \label{fig:articulated_gallery}
\end{figure}

\begin{figure*}[!htbp]
    \centering
    \begin{subfigure}[t]{0.48\textwidth}
        \centering
        \includegraphics[width=\linewidth]{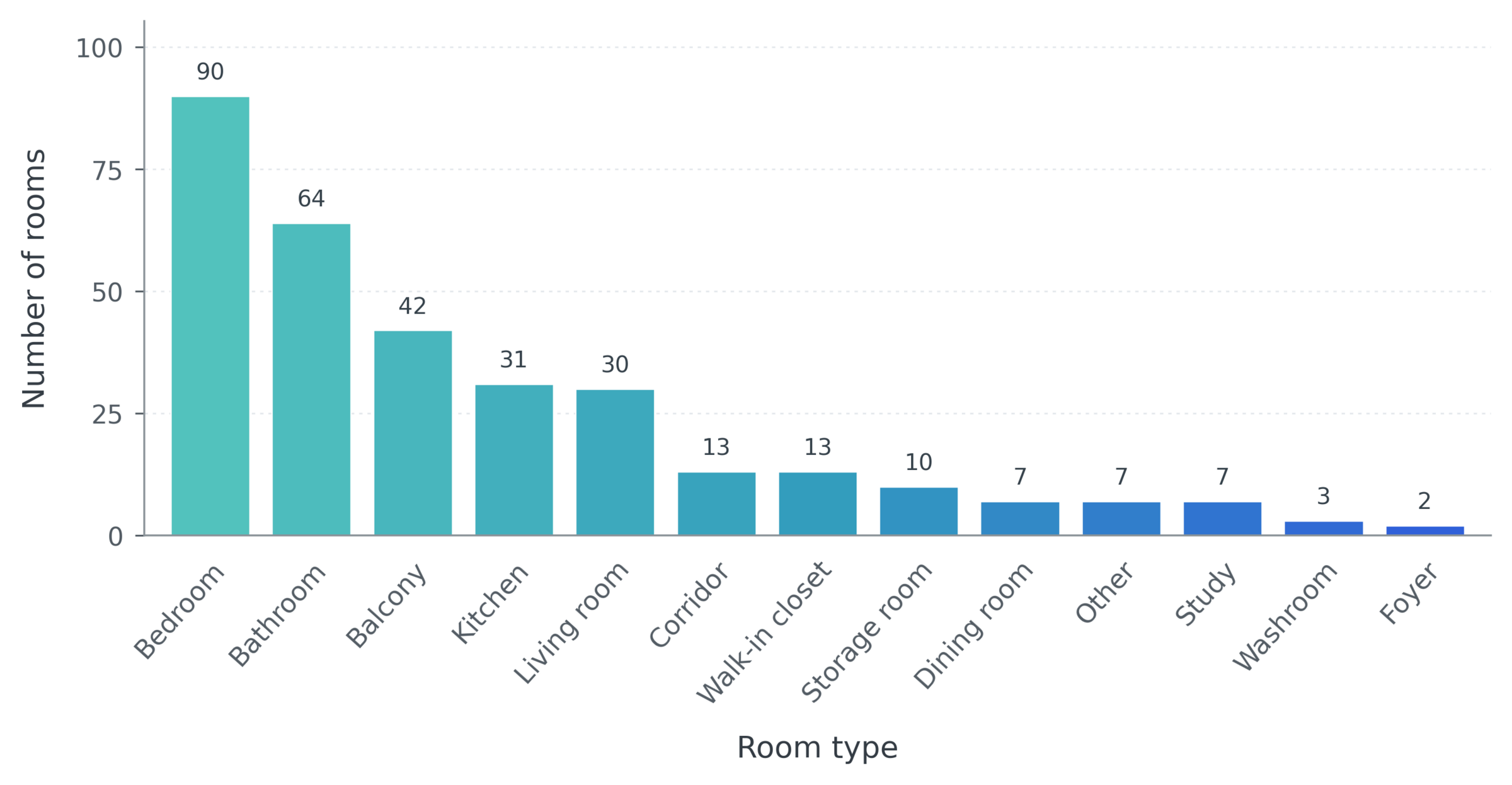}
        \caption{Room-type distribution.}
        \label{fig:room_type_dis}
    \end{subfigure}
    \hfill
    \begin{subfigure}[t]{0.48\textwidth}
        \centering
        \includegraphics[width=\linewidth]{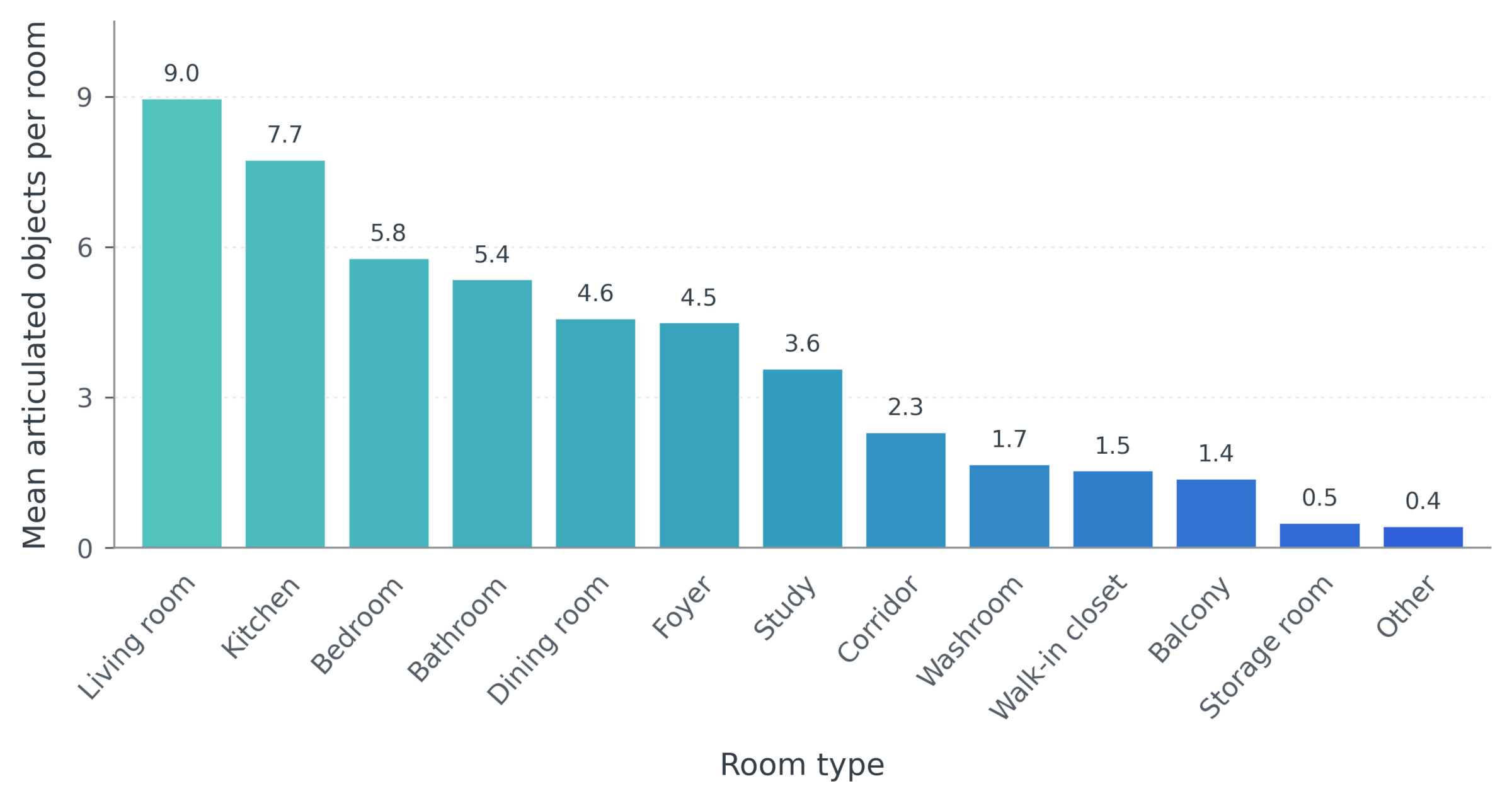}
        \caption{Articulated objects across room types.}
        \label{fig:articulated_dis}
    \end{subfigure}

    \caption{
    \textbf{Statistics of the LIVIN scene collection.}
    We report (a) the distribution of room types and
    (b) the distribution of articulated objects across room types.
    }
    \label{fig:dataset_statistics}
\end{figure*}

We further characterize the \textsc{LIVIN} benchmark through additional dataset statistics and qualitative visualizations. Fig.~\ref{fig:gallery} shows top-down and close-up views of the reconstructed homes, illustrating their layouts and object arrangements. These homes span a diverse set of room types, as shown in Fig.~\ref{fig:room_type_dis}, with bedrooms and bathrooms being the most common, followed by balconies, kitchens, and living rooms. We further report the mean number of articulated objects per room type in Fig.~\ref{fig:articulated_dis}; living rooms and kitchens contain the highest averages, with 9.0 and 7.7 articulated objects per room, respectively, while bedrooms and bathrooms also exhibit substantial articulated content. Representative articulated assets and their movable configurations are visualized in Fig.~\ref{fig:articulated_gallery}. Together, these statistics and visualizations characterize the diversity, object density, real-world correspondence, and interaction-relevant structure of \textsc{LIVIN}.

%% file: sections/appendix_construction.tex
\section{Automated Annotation and Scene Construction}
\label{app:automation}

\subsection{Input Data and Spatial References}
\label{app:inputs}

\begin{figure}[!htbp]
    \centering
    \includegraphics[width=\linewidth]{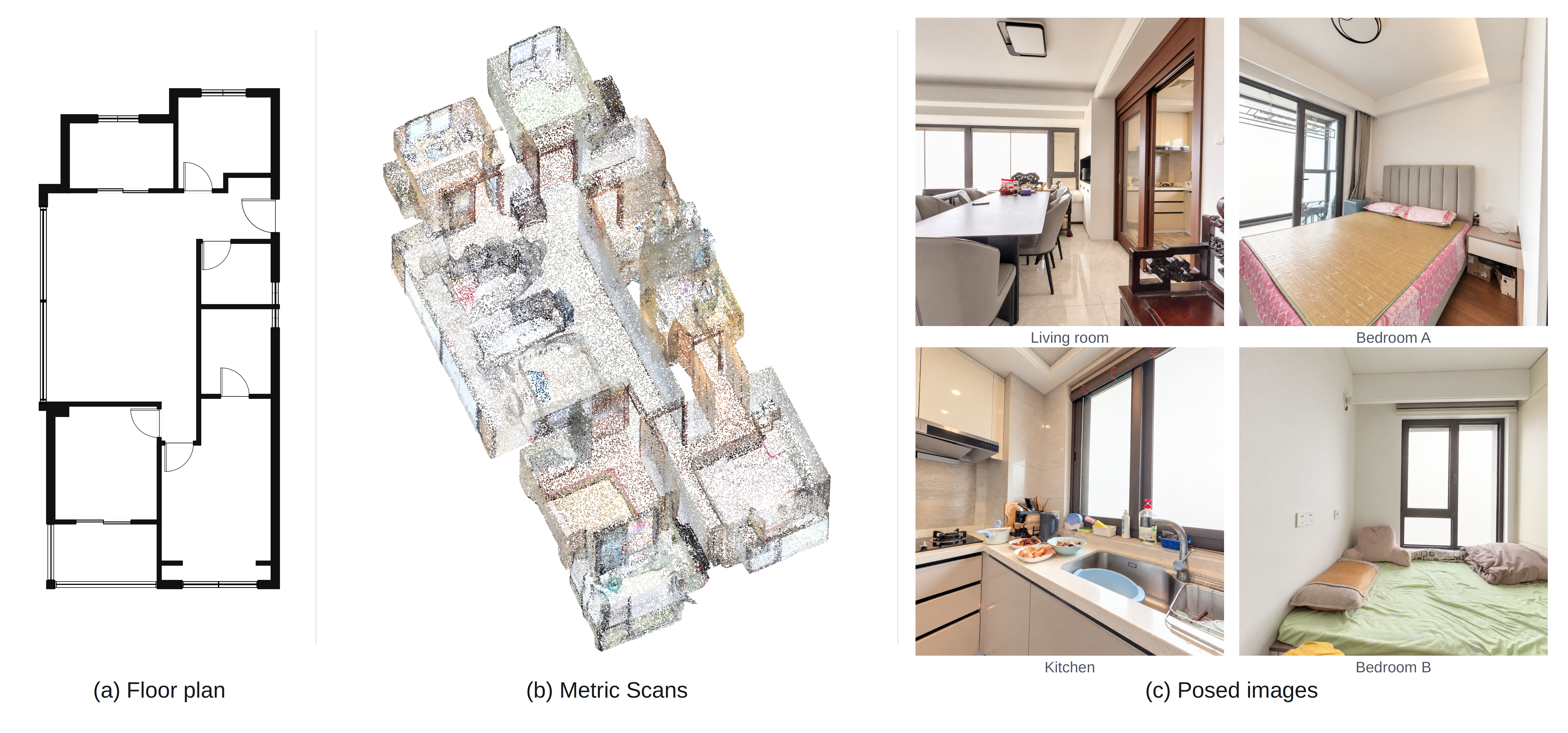}
    \caption{\textbf{Input overview.}
    Each home is represented by a 2D floorplan, a metric scan, and posed multi-view images.}
    \label{fig:construction-inputs}
\end{figure}

Each home is reconstructed from three complementary inputs, illustrated in
Fig.~\ref{fig:construction-inputs}: a floorplan, a metric scan, and posed multi-view images. The floorplan provides room organization and an initial architectural layout, while the metric scan supplies coarse 3D geometry and scale. The posed images provide the detailed visual evidence needed to identify individual objects, recover their appearance, and infer their spatial arrangements. We align these inputs in a common coordinate system before reconstruction. The posed images retain camera calibration and projection information, allowing visual observations to be related to the reconstructed geometry. We also have some structural metadata, including room boundaries and semantic labels, wall geometry and heights, door and window locations and dimensions, and room connectivity, which serve as spatial constraints for constructing the initial architectural structure.

\subsection{Scene-to-Assets Annotation}
\label{app:scene-to-assets}

We create a skill to guide the agent through instance identification, cross-view association, and evidence preparation. Dense or occluded regions are examined using enlarged crops or complementary views. Appearance, landmarks, and spatial evidence are used to match observations with existing entries and remove duplicate instances. 

\paragraph{Semantic and articulation annotations.}
We categorize each instance as articulated or non-articulated, and as large furniture or a small object. The furniture/small-object label provides the agent with a placement-order reference in the later reconstruction stage.  Separate fields record semantic labels, rigid/soft/partially soft composition, and articulation evidence. Each judgment is assigned a confidence level and marked as observed, inferred, or unresolved.

\paragraph{Reference selection and product images.}
The agent selects a clear, sufficiently complete captured view and any needed auxiliary observations while continuing coverage inspection. For non-articulated small objects, the captured view is converted into a product-style reference image using GPT-Image-2~\citep{gptimage2}. The current package requests GPT-Image-2 at medium quality and $1024\times1024$, retaining the original response and delivered image separately. Generated hidden surfaces remain inferences and are checked against capture evidence.

\subsection{Rodin Generation}
\label{app:rodin}

Rodin uses the reference images described above to generate non-articulated 3D assets. Tab.~\ref{tab:replacement-parameters} lists the current integration settings. During subsequent human-guided replacement and refinement, the image hash, prompt, generation parameters, and generation identifier are recorded for each regeneration job, allowing replacement assets to be traced throughout the correction process.

\begin{table}[t]
    \centering
    \caption{
    \textbf{Rodin parameters used for object generation.}
    }
    \label{tab:replacement-parameters}
    \small
    \renewcommand{\arraystretch}{1.12}

    \begin{tabular}{ll}
        \toprule
        \textbf{Parameter}
        & \textbf{Setting}
        \\
        \midrule

        Generator Tier
        & Gen-2.5-High
        \\

        Mesh Mode
        & Raw
        \\

        Quality
        & Low
        \\

        Output
        & GLB
        \\

        Material
        & PBR
        \\

        Texture Mode
        & High
        \\

        Texture Delight
        & Enabled
        \\

        HD Texture
        & Disabled
        \\

        Geometry Instruction Mode
        & Faithful
        \\

        \bottomrule
    \end{tabular}
\end{table}

\subsection{Blender-based Scene Generation}
\label{app:agent-protocol}

We use a coordinating skill to assign room-level tasks to the reconstruction agent, specifying the active stage, input scene, source inventory, accepted base, coordinate conventions, and expected outputs. Tab.~\ref{tab:construction-skills} summarizes these responsibilities. To improve efficiency and reduce generation cost, the agent performs a single feed-forward pass at each stage without iterative self-refinement. Remaining errors are then identified by human reviewers, who provide precise, localized correction instructions to avoid repeated or unnecessary agent revisions.

\paragraph{Initial Structure Generation.}
We obtain the floor elevation from the room metadata and construct an initial 3D architectural structure using the corresponding wall heights and thicknesses. Door and window openings are reserved according to their recorded locations and dimensions. We then align this coarse structure with the metric scan and refine its geometry accordingly. In particular, when the reconstructed ceiling is substantially higher than the ceiling observed in the scan, we lower it and truncate the surrounding walls to the revised height for better geometric consistency.

\paragraph{Modeling in context.}
Articulated objects are generated jointly with their placement so that geometry can adapt to the surrounding scene rather than being positioned only after modeling. For example, cabinets and counters are constructed to fit supporting walls, available footprints, and neighboring structures. The agent also models internal compartments, drawer boxes, handles, and separate fixed and movable parts, with rotation axes, pivots, and sliding directions specified according to the inferred mechanism. To facilitate human verification, we also generate a simple animation for each articulated object to visualize its motion.

\begin{table}[t]
    \centering
    \caption{
    \textbf{Responsibilities of the construction skills.}
    Each skill specifies task instructions and required deliverables for scene construction.
    }
    \label{tab:construction-skills}
    \small
    \renewcommand{\arraystretch}{1.12}

    \begin{tabularx}{\linewidth}{
        @{}
        p{0.18\linewidth}
        >{\raggedright\arraybackslash}X
        @{}
    }
        \toprule
        \textbf{Skill}
        & \textbf{Guidance for the Generated Blender Scene}
        \\
        \midrule

        Coordination
        & Assign stage-specific room tasks, preserve source identities and accepted bases, and collect complete room outputs before publication.
        \\

        Structure
        & Import the initial shell; check cameras and scale; refine walls, openings, recesses, and visible architectural details; reconstruct windows at the reference pose.
        \\

        Structure-Aware Fitting and Articulation
        & Jointly model object geometry, materials, placement, internal structures, and movable parts in the surrounding room context; record articulation mechanisms, support relations, and observed or inferred parameters.
        \\

        Object Placement
        & Import assigned meshes, preserve their parts and materials, place supports before supported objects, ensure collision-free placement, and record root transforms and support relations.
        \\

        \bottomrule
    \end{tabularx}
\end{table}

\subsection{Construction Ablations}
\label{app:construction-ablations}

\begin{figure}[!htbp]
    \centering
    \includegraphics[width=\linewidth]{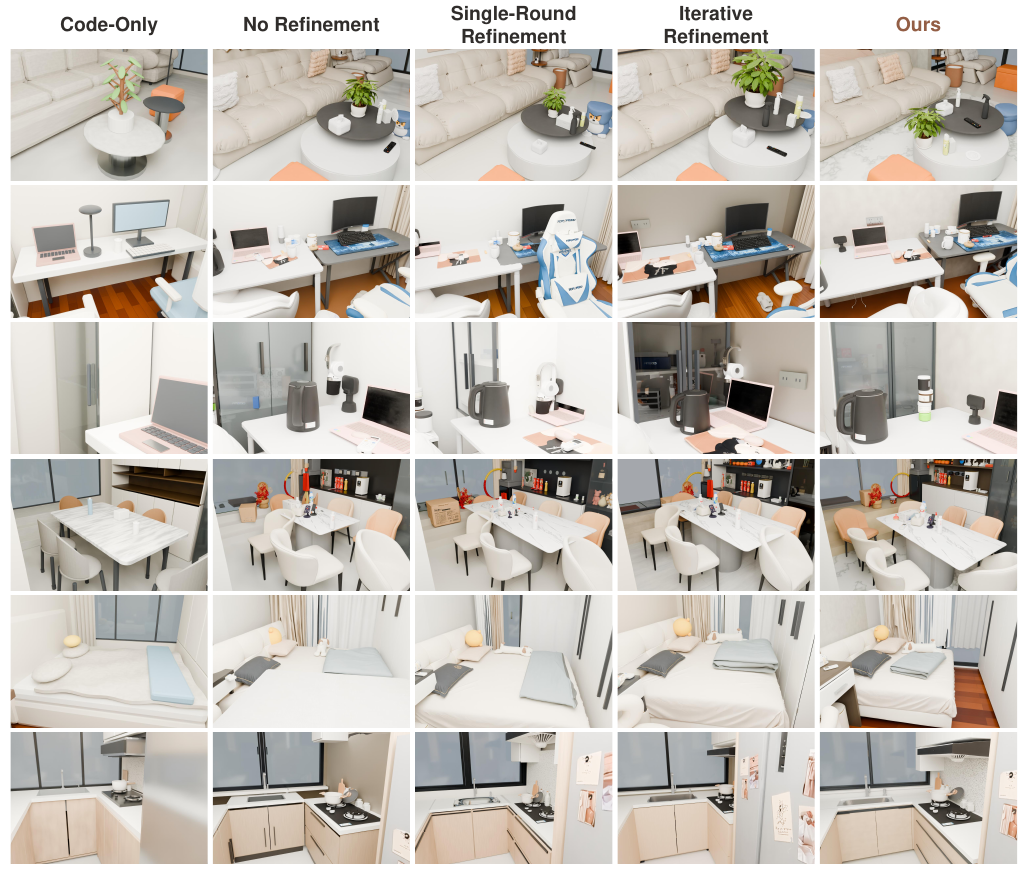}
    \caption{
    \textbf{Construction ablation.}
    Qualitative comparison of Code-Only, Single-Round Refinement, Iterative Refinement, No Refinement, and our human-guided construction strategy.
    }
    \label{fig:ablation}
\end{figure}

We ablate the construction and correction strategy using five settings, with some qualitative results shown in Fig.~\ref{fig:ablation}. \textbf{Code-Only} reconstructs the entire scene directly from floorplans, posed images, and metric scans using Blender code, followed by up to five rounds of agent self-refinement. \textbf{No Refinement}, \textbf{Single-Round Refinement}, and \textbf{Iterative Refinement} use the same staged initialization with fixed coarse structural meshes, detection results, and Rodin assets, but apply zero, one, and up to three rounds of autonomous refinement, respectively. To assess the efficiency of the reconstruction process, we also record token consumption. \textbf{Single-Round Refinement} and \textbf{Iterative Refinement} each require over 100 million tokens, compared with just over 70 million for our workflow. The results show that our workflow improves physical plausibility and mesh quality while reducing inter-object penetrations. These gains arise from both our generation strategies and human-in-the-loop verification. Blender-based construction preserves articulated structure and kinematics, while Rodin provides high-fidelity geometry for non-articulated objects. Human review further improves geometry and placement accuracy, while avoiding redundant agent-side checking and ineffective iterative refinement, leading to a better balance between reconstruction quality and computational cost.

%% file: sections/appendix_review.tex
\section{Human Review and Verification}
\label{app:annotation}

\subsection{Review and Correction Workflow}
\label{app:detection-review}

Before scene reconstruction, reviewers inspect the instance inventory against the source observations, with particular attention to low-confidence or unresolved cases from the previous stage. They resolve uncertain annotations, add missing instances, remove duplicates, and revise semantic labels or room assignments when needed. Reviewers can also update the reference image used for downstream asset generation. When necessary, they provide a text prompt to GPT-Image-2 to regenerate a cleaner product-style reference while preserving the target object's identity and appearance.

For 3D reconstruction, reviewers first inspect and correct the generated architectural structure. Object generation and placement are then performed automatically, after which the resulting object-level reconstruction is reviewed in a second stage. Articulated and non-articulated objects are corrected using different strategies, as summarized in Tab.~\ref{tab:annotation-protocol}.

\begin{table}[htbp]
    \centering
    \caption{
    \textbf{3D reconstruction review process.}
    Each phase is reviewed for its corresponding reconstruction aspects, followed by targeted correction and completion.
    }
    \label{tab:annotation-protocol}
    \small
    \renewcommand{\arraystretch}{1.12}

    \begin{tabularx}{\linewidth}{
        @{}
        p{0.15\linewidth}
        >{\raggedright\arraybackslash}X
        >{\raggedright\arraybackslash}X
        @{}
    }
        \toprule
        \textbf{Phase}
        & \textbf{Inspection Aspects}
        & \textbf{Correction and Completion}
        \\
        \midrule

        Structure
        & Room boundaries, wall connections, openings, scale, and alignment with source observations.
        & Select the relevant structural element and provide localized correction feedback. Each room is explicitly confirmed before proceeding.
        \\

        Articulated Objects
        & Geometry, dimensions, materials, placement, structural fit, and articulation.
        & Select the relevant object or part for correction, adjust its pose, and resolve inter-object penetration.
        \\

        Non-Articulated Objects
        & Appearance, geometry, placement, support, and relations to surrounding objects.
        & Adjust object poses, resolve inter-object penetration, or regenerate defective assets.
        \\

        \bottomrule
    \end{tabularx}
\end{table}

Architectural structures and articulated objects permit at most two agent repair rounds based on human feedback. Each round starts from a copy of the saved human-adjusted Blender scene, addresses the latest review feedback, and returns all rooms for another round of inspection. Non-articulated objects are corrected through Rodin regeneration or manual pose adjustment, without additional agent repair rounds.

\subsection{Reconstruction Review Interface}
\label{app:review-interface}

\begin{figure}[ht]
    \centering
    \includegraphics[width=\linewidth]{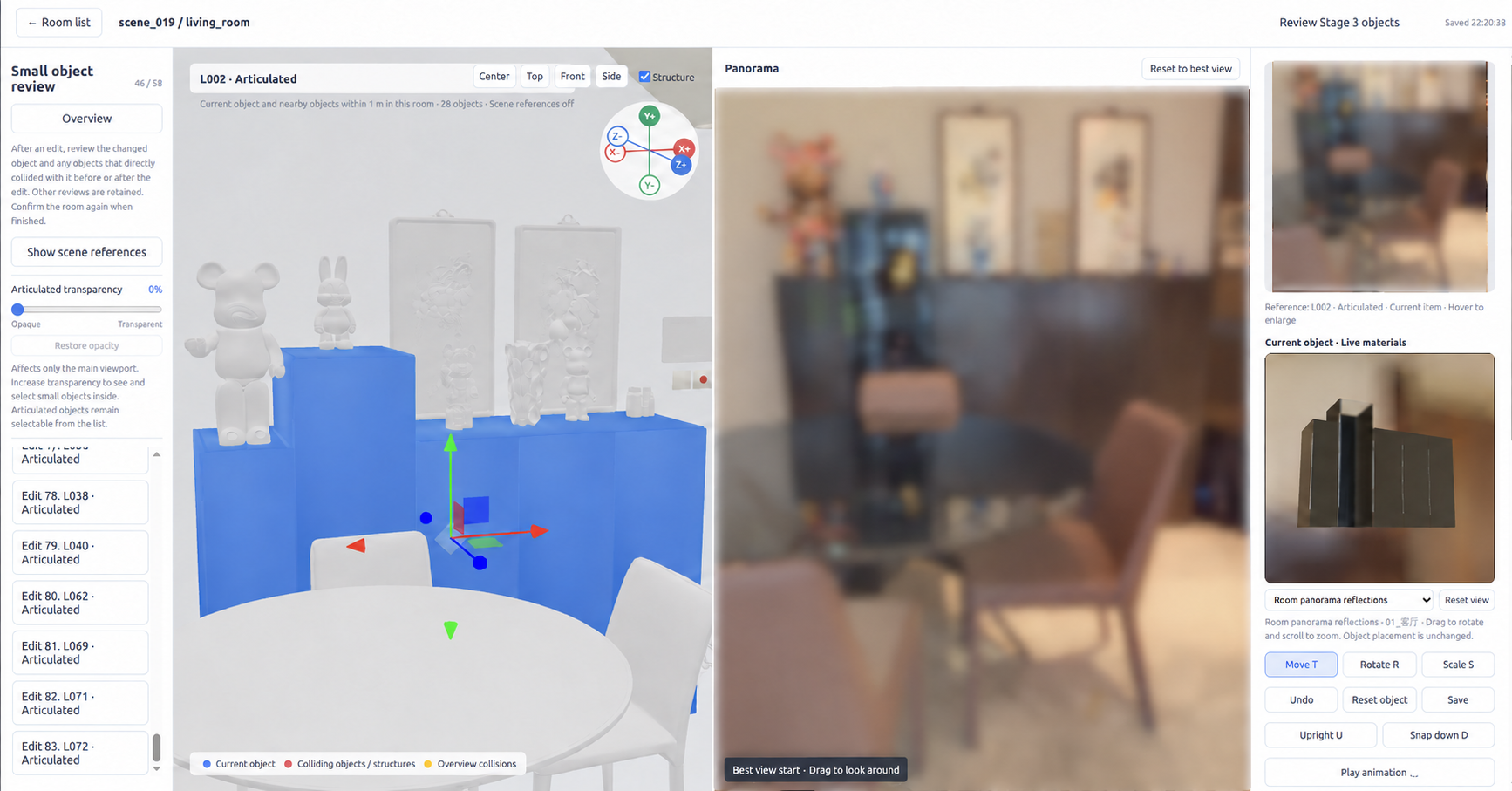}
    \caption{
        \textbf{Snapshot of the human review interface.}
        The reconstructed scene is shown on the left, the corresponding posed reference image on the right, and an additional panel on the far right displays object material information for inspection.
    }
    \label{fig:human-review-interface}
\end{figure}

\paragraph{Model--observation comparison.}
The interface places the reconstructed model beside the captured posed images, as shown in Fig.~\ref{fig:human-review-interface}. Selecting an asset initializes the best captured view at its representative observation point and viewing direction. Reviewers can switch among posed images and zoom in for closer inspection.

\paragraph{Geometric context and collision feedback.}
Scenes in \textsc{LIVIN} contain many high-resolution object meshes, making it memory-intensive to load the entire scene into the review interface at once. We therefore display the selected object together with nearby objects within 1\,m, while always including its detected collision partners. The selected object is highlighted in blue and collision partners in red, allowing reviewers to inspect local geometric relationships without loading all scene assets simultaneously. Collision feedback is computed on reference-pose meshes using bounding-box filtering followed by BVH triangle intersection.

\paragraph{Pose editing and snapping.}
For placement errors that do not require asset regeneration, reviewers can directly correct object poses through translation, rotation and uniform scaling. However, for floating objects or objects that must closely align with supporting surfaces, manual pose adjustment alone cannot reliably establish precise contact. We therefore introduce axis-based snapping, which translates an object along a user-specified world axis while preserving its rotation and scale. Candidate contact surfaces are first filtered using bounding boxes, after which BVH-accelerated mesh intersection tests search for the first contact and a binary search refines the contact position. The final position is offset by 2\,mm from the detected contact surface along the snapping direction, providing a small safety margin against mesh penetration. This allows supported objects to settle onto floors or furniture and wall-mounted or switch-like structures to align closely with architectural surfaces.

\paragraph{Animation and material inspection.}
To better visualize the articulation of articulated objects, the interface allows reviewers to play simple motion previews of their movable parts. Rigid-part animation previews expose the authored motion relative to the reference geometry and retain explicit statuses for absent or unsupported motion. As shown in the far-right panel of Fig.~\ref{fig:human-review-interface}, an on-demand PBR preview allows reviewers to inspect object textures, transparency, reflections, and material channels under controlled illumination. 

 \paragraph{Text feedback and bounding-box control.}
The interface provides a dedicated text box whose purpose depends on the selected element: reviewers enter correction instructions for the GPT agent when reviewing room structures or articulated objects, and generation prompts for Rodin when reviewing non-articulated objects. For Rodin generation, reviewers also can optionally adjust a 3D bounding box through axis dragging or numerical inputs to control the object's width, height, and depth proportions. Bounding-box control helps Rodin reconstruct object geometry more accurately by constraining its overall proportions.

%% file: sections/appendix_reconstruction.tex
\section{3D Scene Reconstruction}
\label{app:3d_reconstruction}

\paragraph{Main evaluation protocol.}
CD-S measures the symmetric mean squared nearest-neighbor surface distance,
in $\mathrm{m}^{2}$.
The main-table F-Score-S uses a fixed distance threshold of $0.05\,\mathrm{m}$,
while object-level F1 uses a 3D box-matching threshold of $0.25$.
F1 is computed from TP, FP, and FN pooled across views; the remaining
main-table metrics are averaged over views.

For appearance evaluation, predictions are independently rendered using
common cameras, lighting, and black backgrounds, without GT-mask clipping.
GT images retain brightness-adjusted RGB values within the selected target
masks and are black elsewhere.
PSNR, SSIM, and LPIPS are computed over the full images.
ShapeR is excluded from appearance evaluation because it produces
untextured geometry.

\paragraph{Clutter-focused evaluation.}

\begin{figure*}[t]
    \centering
    \begin{minipage}[t]{0.49\textwidth}
        \centering
        \includegraphics[width=\linewidth]{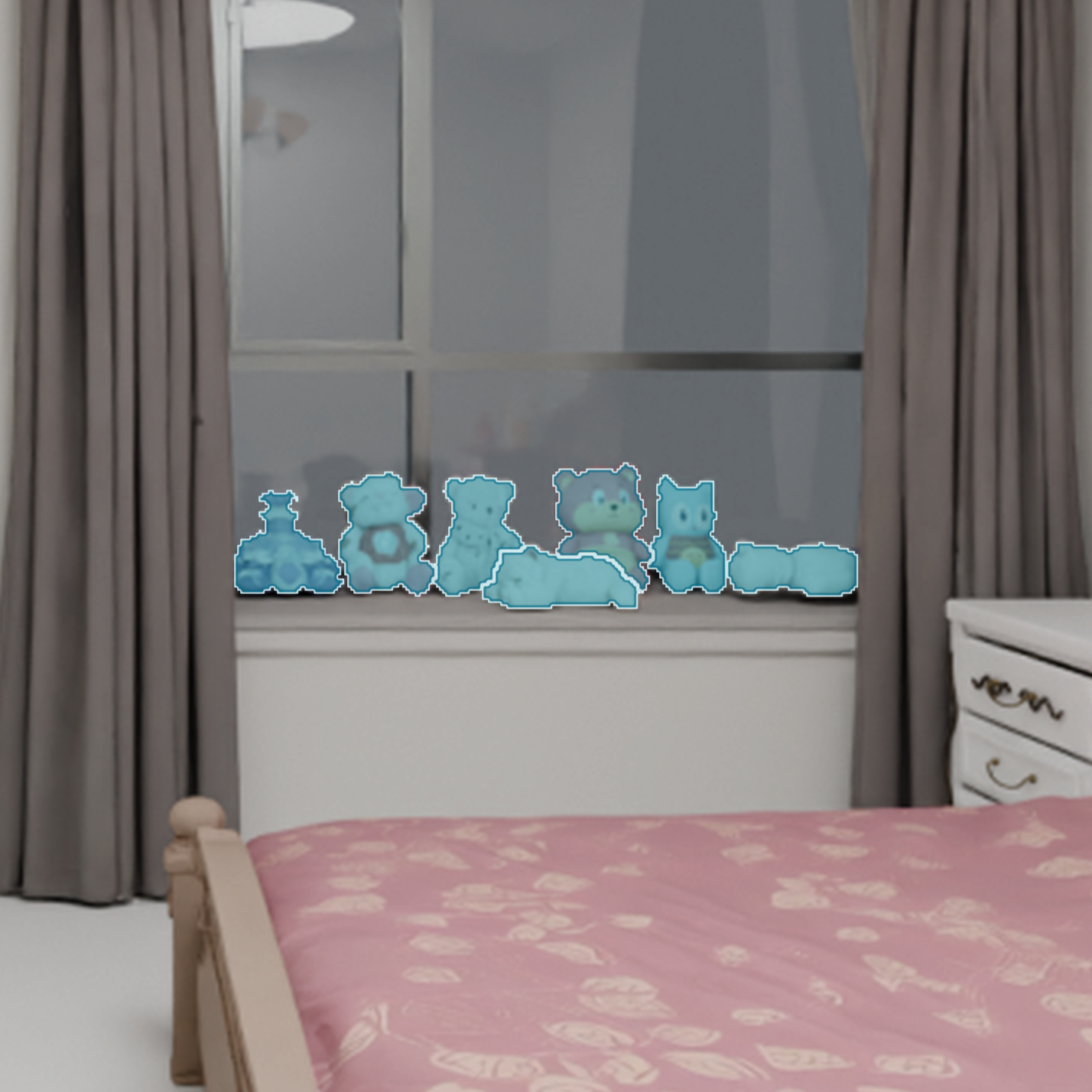}
        \par\vspace{0.5em}
        \includegraphics[width=\linewidth]{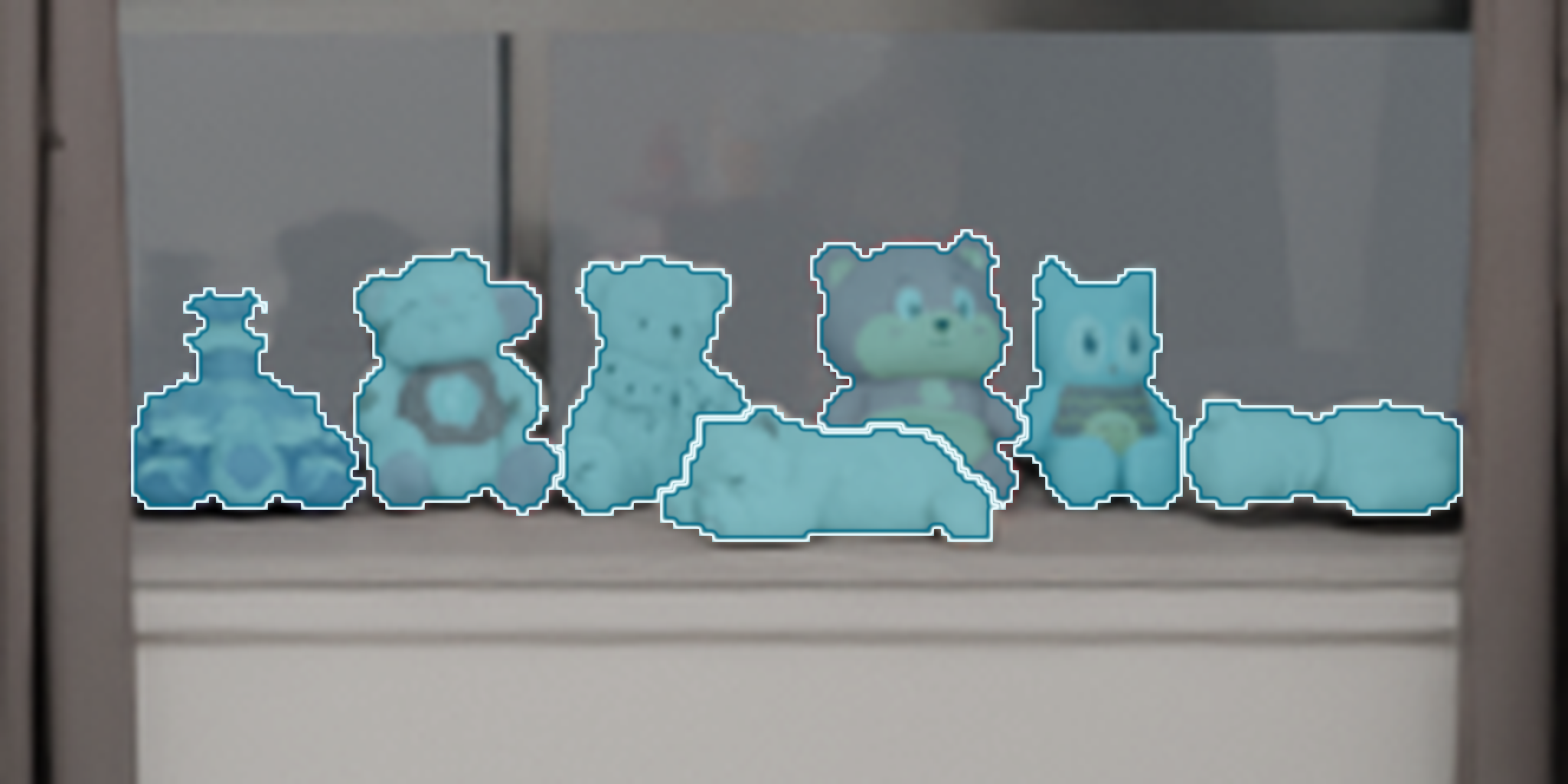}
    \end{minipage}%
    \hfill
    \begin{minipage}[t]{0.49\textwidth}
        \centering
        \includegraphics[width=\linewidth]{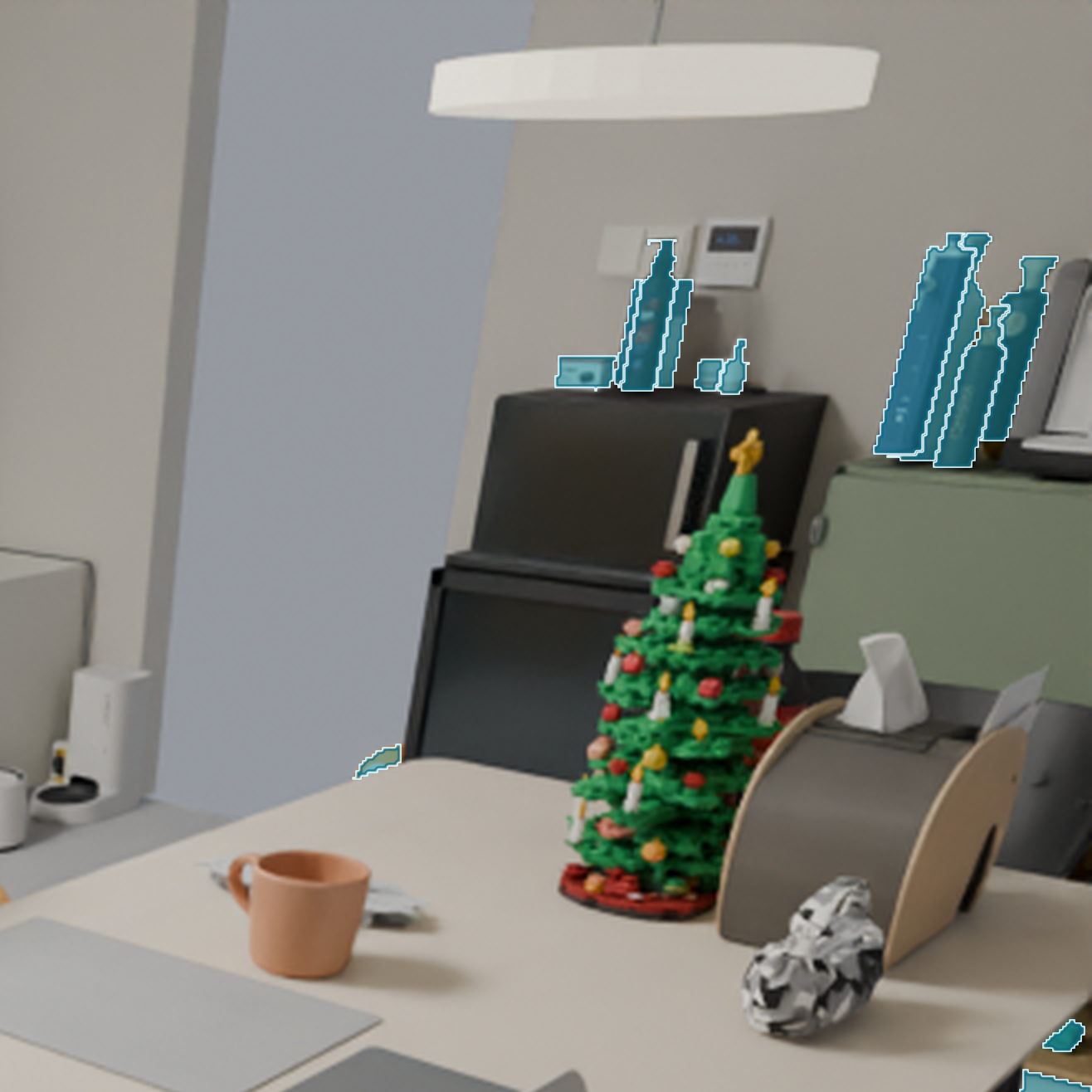}
        \par\vspace{0.5em}
        \includegraphics[width=\linewidth]{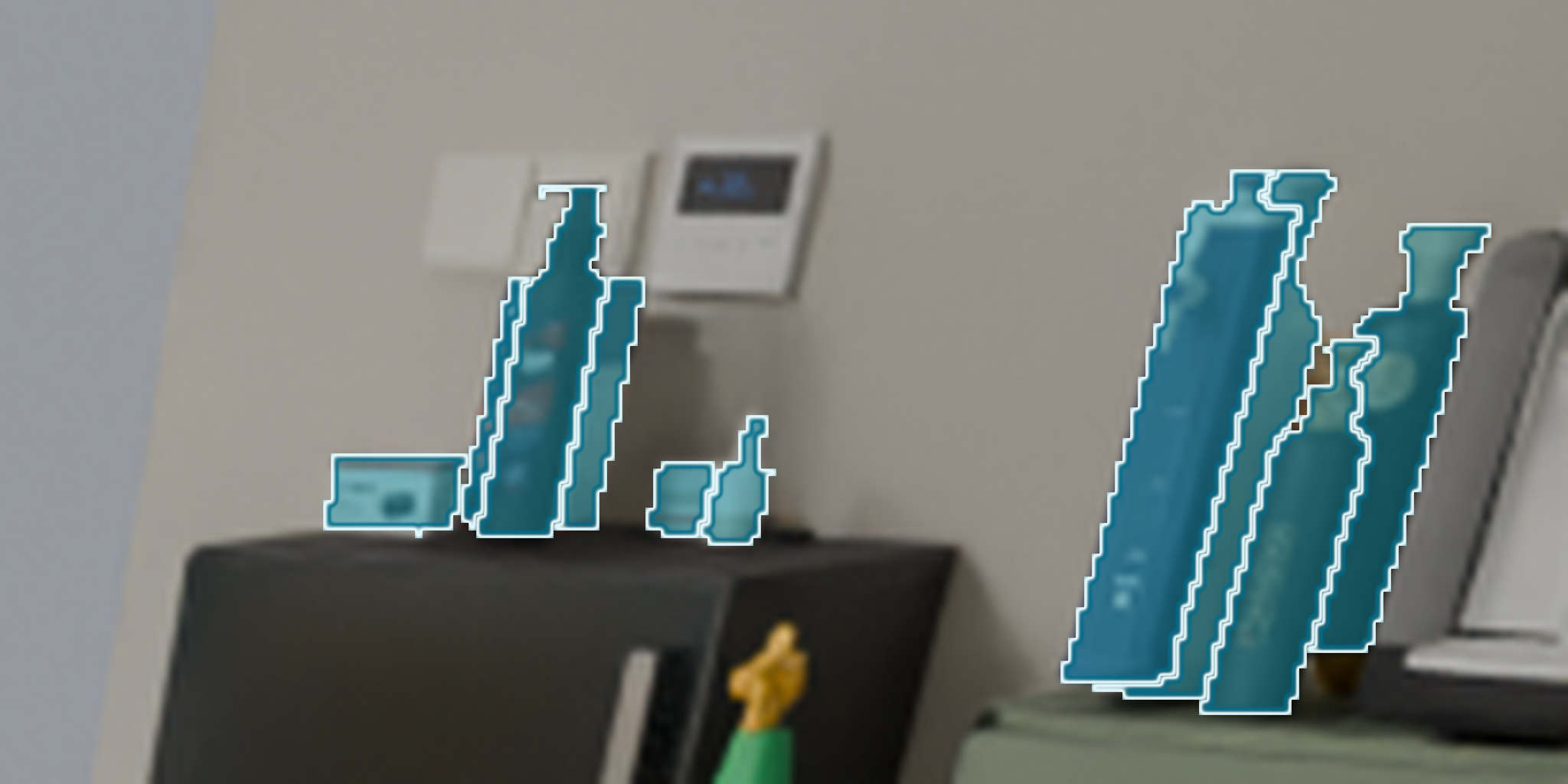}
    \end{minipage}
    \caption{\textbf{Selected small clutter objects from the \textsc{LIVIN} test set.} Top: full-scene views. Bottom: corresponding close-ups. Highlighted objects belong to the common evaluation subset used by all methods.}
    \label{fig:clutter_selection}
\end{figure*}

We further compare different reconstruction methods in terms of small-object coverage and geometric fidelity. A common evaluation subset is drawn from the main 128-view \textsc{LIVIN} test set for all methods. We first use GT name filtering to select clutter-related objects, and then apply physical and projected size constraints to select small targets. A minimum projected-size threshold further removes barely visible instances. The resulting subset contains $306$ object--view records covering $287$ distinct objects across $76$ views. Fig.~\ref{fig:clutter_selection} shows examples of the selected objects.

We report output coverage, 3D box recall, and aggregate surface F-score in Tab.~\ref{tab:clutter_reconstruction}. ShapeR performs best on this subset, achieving full output coverage together with substantially higher box recall and surface F-score. Fire3D also achieves strong geometric accuracy, but its lower coverage indicates that some objects are not reconstructed at all. Gen3DSR exhibits the opposite limitation: its small-object filtering leaves more than half of the selected records without output, making coverage the dominant bottleneck. Most methods show a marked drop in 3D box recall under cluttered small-object conditions, and all methods still leave substantial room for improvement. These results demonstrate that \textsc{LIVIN} effectively exposes current limitations in small-object reconstruction and provides a meaningful benchmark for comparing reconstruction methods in challenging cluttered environments.

\begin{table}[t]
    \centering
    \caption{
    \textbf{Output coverage and reconstruction accuracy on the clutter-like proxy subset of \textsc{LIVIN}.}
    Output coverage reports the number and percentage of generated objects, while 3D box recall and $\overline{F}_{\mathrm{obj}}$ evaluate object-level reconstruction accuracy.
    }
    \label{tab:clutter_reconstruction}
    \small
    \setlength{\tabcolsep}{6pt}
    \renewcommand{\arraystretch}{1.12}

    \begin{tabular}{lccc}
        \toprule
        \textbf{Method}
        & \textbf{Output Coverage} $\uparrow$
        & \textbf{3D Box Recall} $\uparrow$
        & \textbf{$\overline{F}_{\mathrm{obj}}$} $\uparrow$
        \\
        \midrule

        Gen3DSR~\citep{Ardelean2025Gen3DSR}
        & 132/306 (43.14\%)
        & 34.97\%
        & 0.2586
        \\

        RecGen~\citep{zadaianchuk2026reconstruction}
        & \textbf{306/306 (100.00\%)}
        & 42.48\%
        & 0.3558
        \\

        SAM3D~\citep{sam3d}
        & \textbf{306/306 (100.00\%)}
        & 53.59\%
        & 0.3686
        \\

        Fire3D~\citep{xia2026fire3d}
        & 292/306 (95.42\%)
        & 72.55\%
        & 0.5774
        \\

        ShapeR~\citep{Siddiqui_2026_CVPR}
        & \textbf{306/306 (100.00\%)}
        & \textbf{85.95\%}
        & \textbf{0.7510}
        \\

        \bottomrule
    \end{tabular}
\end{table}

%% file: sections/appendix_navigation.tex
\section{Navigation Evaluation Details}
\label{app:navigation}

\subsection{Implementation Details}
All GR00T-based systems use the same whole-body control backend.
GPT-6 Astra~\citep{gpt6}, configured with high reasoning effort, observes RGB images from the mounted head camera on demand and maintains within-episode interaction history. It outputs body-frame planar velocity, yaw rate, and execution duration, alternating between action execution and new observations as needed.

OmniNav~\citep{xue2026omninav} takes front, left, and right RGB views together with visual history and predicts planar waypoints and relative headings, which we convert into local GR00T commands. DualVLN~\citep{wei2026ground} takes head-camera RGB observations and visual history and follows its original System-2 planning and System-1 metric trajectory prediction pipeline. Our adapter executes each predicted local subgoal for $1\,\mathrm{s}$ before acquiring a new observation and replanning; the subgoal need not be reached within that interval.

\subsection{Experimental Setting}

All systems control the same simulated Unitree G1 humanoid on a shared set of 45 navigation tasks and environments. For each task, the start and goal locations are manually selected, and a reference path is computed with A* search on the scene NavMesh. We generate an initial route instruction from the reference path, room layout, and object semantics by selecting landmarks and route events, then converting their ordered motion and spatial relations into natural language with rule-based templates. A video following the reference path is rendered for human review, and annotators revise the instruction to ensure route--instruction consistency, reduce ambiguity among alternative paths, and make the destination clearly identifiable. Each system is evaluated three times on every task, yielding 135 episodes per system. Each episode begins with a $3\,\mathrm{s}$ stabilization phase, after which navigation is limited to $240\,\mathrm{s}$ of simulation time and $900\,\mathrm{s}$ of total runtime. The main evaluation metrics are introduced below.

\paragraph{Navigation error (NE).}
For execution $r$ of task $i$, navigation error is
\begin{equation}
\mathrm{NE}_{ir}
=
d_{\mathrm{nav}}
\left(
\boldsymbol{p}^{\,\mathrm{final}}_{ir},
\boldsymbol{g}_i
\right) m,
\end{equation}
where $\boldsymbol{p}^{\,\mathrm{final}}_{ir}$ is the final planar robot
position and $\boldsymbol{g}_i$ is the destination.
The navigable distance $d_{\mathrm{nav}}$ is computed after projection onto
the evaluation NavMesh using triangle-graph A* search followed by funnel
refinement.

\paragraph{Success rate (SR).}
An execution is successful only if the system explicitly issues
\textsc{Stop} and the final navigable distance to the destination is at most
$\tau=1\,\mathrm{m}$:
\begin{equation}
S_{ir}
=
\mathbb{I}
\left[
\text{\textsc{Stop}}
\;\land\;
\mathrm{NE}_{ir}\leq\tau
\right].
\end{equation}
We use a relatively strict $1 \mathrm{m}$ threshold because \textsc{LIVIN} preserves the dense object arrangements of real lived-in homes, where subsequent interaction often requires the robot to reach the immediate vicinity of a specific object or interaction region. SR is the aggregated success indicator and is reported as a percentage.

\paragraph{Success weighted by path length (SPL).}
For each execution,
\begin{equation}
\mathrm{SPL}_{ir}
=
S_{ir}
\frac{L_i}{\max(L_i,P_{ir})},
\end{equation}
where $L_i$ is the start-to-destination reference path length on the
evaluation NavMesh and $P_{ir}$ is the accumulated planar trajectory length
of the robot base after stabilization.
Aggregated SPL is reported as a percentage.

\paragraph{Collision counts.}
All systems use the same MuJoCo~\citep{mujoco} collision geometry at a voxel resolution of $2.5\,\mathrm{cm}$. Contacts involving multiple robot links at the same time are counted as one collision event. Contact intervals separated by less than $0.1\,\mathrm{s}$ are merged, and a new event is counted only after all eligible contacts have ended for at least $0.1\,\mathrm{s}$.

\subsection{Further Analysis of Navigation Performance}
\label{app:navigation_analysis}

\paragraph{Failure case analysis.}
To better understand the differences between navigation methods, we further analyze the observed failure patterns. We group failed episodes according to the execution evidence into unrecovered blockage, falls, goal-completion mismatch, and other failures, such as repetitive turning and execution stalls. We also report the mean navigation error (NE) over failed episodes to characterize how far the robot remains from the goal when navigation fails. The statistics are summarized in Tab.~\ref{tab:failure_analysis}.

\begin{table}[t]
    \centering
    \caption{
    \textbf{Failure case analysis of navigation methods on \textsc{LIVIN}.}
    Failure categories are reported as percentages of failed episodes for each method.
    Failed NE denotes the mean navigation error over failed episodes.
    }
    \label{tab:failure_analysis}
    \small
    \setlength{\tabcolsep}{2.5pt}
    \renewcommand{\arraystretch}{1.12}

    \begin{tabular*}{\linewidth}{
        @{\extracolsep{\fill}}
        lccccc
        @{}
    }
        \toprule

        \textbf{Method}
        & \shortstack{\textbf{Unrecovered}\\\textbf{Blockage (\%)}}
        & \shortstack{\textbf{Fall}\\\textbf{(\%)}}
        & \shortstack{\textbf{Goal-Completion}\\\textbf{Mismatch (\%)}}
        & \shortstack{\textbf{Other}\\\textbf{Failures (\%)}}
        & \shortstack{\textbf{Failed NE}\\\textbf{(m)} $\downarrow$}
        \\

        \midrule

        GPT-6 Astra~\citep{gpt6}
        & 1.27
        & 17.72
        & 77.22
        & 3.80
        & \textbf{3.15}
        \\

        OmniNav~\citep{xue2026omninav}
        & 69.05
        & 2.38
        & 15.08
        & 13.49
        & 7.12
        \\

        DualVLN~\citep{wei2026ground}
        & 14.73
        & 0.00
        & 71.32
        & 13.95
        & 6.22
        \\

        \bottomrule
    \end{tabular*}
\end{table}

The failure statistics further support the observations in Sec.~\ref{sec:navigation}. For OmniNav, unrecovered blockage is the dominant failure mode, accounting for 69.05\% of failed episodes, while DualVLN shows a much lower blockage rate  but a large fraction of goal-completion mismatches. GPT-6 Astra is also dominated by goal-completion mismatch, but its failed-episode NE is substantially lower than that of DualVLN, suggesting that many failures occur near the destination and are often caused by inaccurate final stopping-region judgments than by large route deviations.

\begin{figure}[ht]
    \centering
    \includegraphics[width=\linewidth]{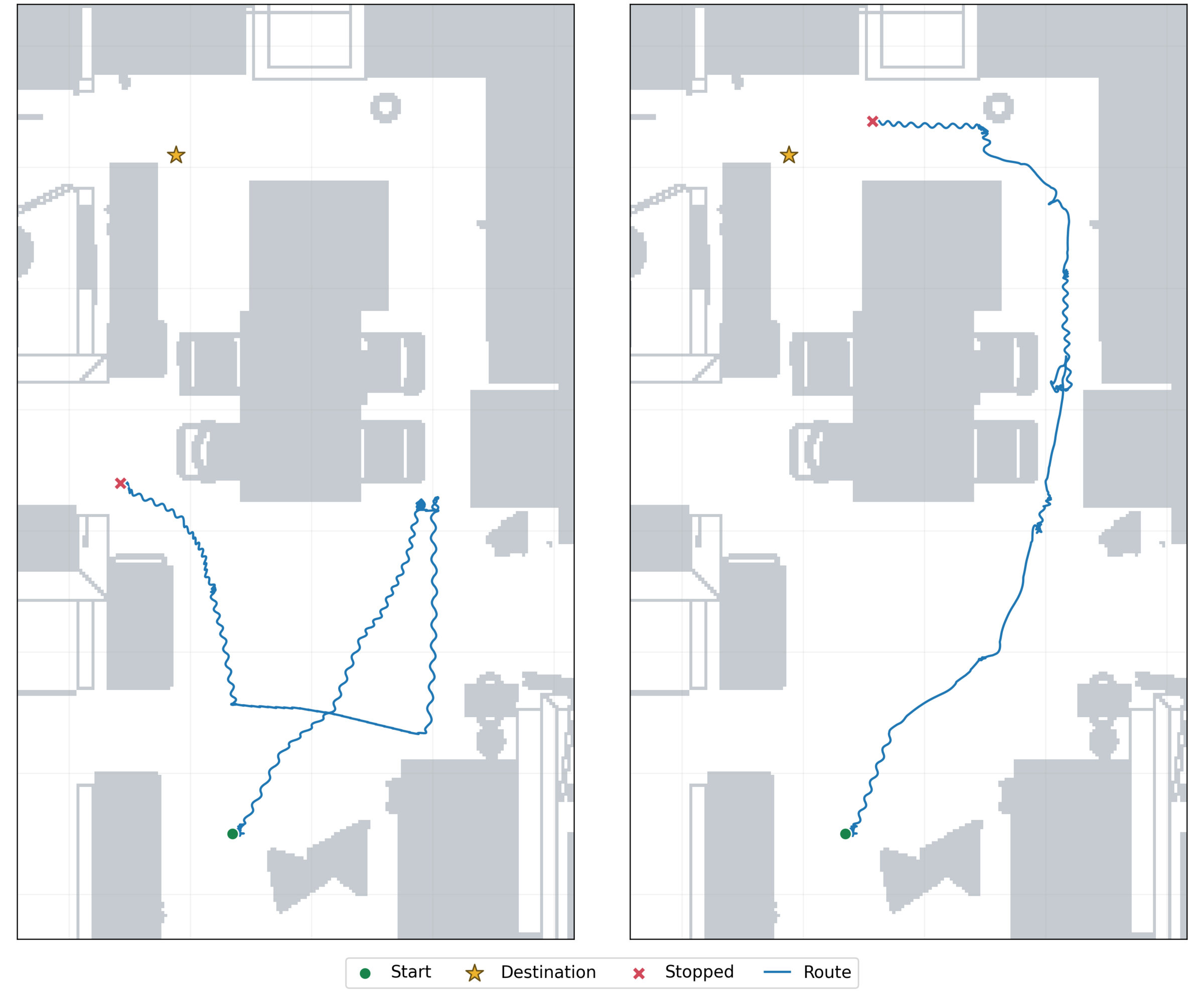}
    \caption{
        \textbf{Navigation trajectories with GR00T and the mixed execution setting.}
    Left: GPT-6 Astra+GR00T fails to pass the narrow passage on the right side of the dining room.
    Right: GPT-6 Astra+Mixed uses CAT for local traversal, allowing the robot to pass through the constrained right-side passage and continue toward the goal.
    }
    \label{fig:mixed_comparision}
\end{figure}

\paragraph{Analysis of the mixed setting.}
We further explore whether introducing an obstacle-avoidance strategy can improve navigation in constrained household spaces. To this end, we use CAT~\citep{CAT}, a collision-free humanoid traversal policy for cluttered environments. During navigation, GPT-6 Astra selects between GR00T and CAT based on the current context, allowing it to switch to CAT when additional obstacle-avoidance capability is needed. This mixed strategy can help in tasks that require the robot to pass through narrow or cluttered spaces, as shown in Fig.~\ref{fig:mixed_comparision}. However, its overall success rate is 34.81\%, which is lower than that of GPT-6 Astra+GR00T. We find that, when CAT is selected for local traversal, the robot often walks sideways, with its body orientation misaligned with the direction of travel. This can make the head-mounted RGB observations seen by the high-level planner inconsistent with the robot's actual motion direction, which may affect subsequent instruction following and turning decisions. 

%% file: sections/appendix_loco-manipulation.tex
\section{Loco-Manipulation Evaluation Details}
\label{app:loco-protocol}

\subsection{Task Inventory}

The benchmark comprises 35 tasks across 30 homes, grouped into object grasping and placement, object pushing and pulling, articulated-object manipulation, and long-horizon tasks with dependent subgoals. Tab.~\ref{tab:loco-tasks} lists the task instructions.

\begin{table}[!htbp]
\centering
\footnotesize
\caption{\textbf{Loco-manipulation tasks.} Instructions are condensed from the task definitions. }
\label{tab:loco-tasks}
\renewcommand{\arraystretch}{1.05}
\begin{tabularx}{\linewidth}{@{}l>{\raggedright\arraybackslash}X@{}}
\toprule
\multicolumn{2}{@{}l}{\textit{Object grasping and placement (15 tasks)}} \\
\midrule
LM1  & Move a saucer from the placemat to an empty spot on the table. \\
LM2  & Grasp and lift a small green cream jar and keep holding it. \\
LM3  & Grasp and lift a handheld massager and hold it stably. \\
LM4  & Grasp and lift a gray insulated cup and keep holding it. \\
LM5  & Place the wine glass on the table to the right of the placemat. \\
LM6  & Lift a toy elephant with both hands and keep holding it. \\
LM7  & Place a cup on the counter to the left of a chocolate jar. \\
LM8  & Move a bottle inward, away from the table edge. \\
LM9  & Grasp and lift a fitness ring and keep holding it. \\
LM10 & Place an oval box on a coaster. \\
LM11 & Move a water bottle to the counter left of the water dispenser. \\
LM12 & Place a remote control on the coffee table to the left of a small red box. \\
LM13 & Lift a tissue box with both hands and keep holding it. \\
LM14 & Lift a vacuum flask off its support with both hands and hold it stably. \\
LM15 & Grasp and lift a fitness ring from the floor and keep holding it. \\
\midrule
\multicolumn{2}{@{}l}{\textit{Articulated-object manipulation (7 tasks)}} \\
\midrule
AM1 & Pull a kitchen base-cabinet door open by its handle to allow access to the interior. \\
AM2 & Open an oven door and then close it. \\
AM3 & Pull out the upper drawer of an under-counter steriliser cabinet far enough to access its contents, and keep it open. \\
AM4 & Close the open upper door of a refrigerator. \\
AM5 & Open a countertop microwave door to allow access to the interior, and keep it open. \\
AM6 & Close both doors at the entrance of a walk-in closet. \\
AM7 & Pull the leftmost under-window kitchen base-cabinet door open by its handle to allow access to the interior. \\
\midrule
\multicolumn{2}{@{}l}{\textit{Object pushing and pulling (5 tasks)}} \\
\midrule
PC1 & Pull a chair backward into the designated target area. \\
PC2 & Move a chair forward into the designated target area. \\
PC3 & Move a study chair to clear the room exit. \\
PC4 & Move the chair beside the dining table aside to make room for people to walk through. \\
PC5 & Push the gray baby stroller forward. \\
\midrule
\multicolumn{2}{@{}l}{\textit{Long-horizon tasks (8 tasks)}} \\
\midrule
LH1 & Move two blocking bottles aside, then lift the blue-capped water bottle and keep holding it. \\
LH2 & Move a pump bottle from the media console below the television to the coffee table. \\
LH3 & Pick up a detergent bottle and place it upright on the kitchen counter beside the sink, then place a small bowl inside the sink. \\
LH4 & Open a kitchen base-cabinet door and place the air fryer from the countertop inside. \\
LH5 & Carry an egg from the kitchen counter to the dining table and set it down beside the tissue box. \\
LH6 & Open the closet door, enter the walk-in closet, and touch the black cloth with one hand. \\
LH7 & Carry a cushion from the living-room sofa to the bed in a bedroom. \\
LH8 & Carry a metal thermos from the bedside table to the living-room coffee table and set it down upright. \\
\bottomrule
\end{tabularx}
\end{table}

\subsection{Simulation and Embodiment}

All tasks use a simulated Unitree G1 humanoid with 29 body joints (12 leg, 3 waist, and 14 arm joints) and two 7-DoF Dex3 hands. Following SIMPLE~\citep{wei2026simple}, we use MuJoCo for physics simulation and Isaac Sim for rendering, with a physics time step of 5\,ms. The simulation pauses during model inference and resumes for action execution.
\subsection{Evaluated Methods }

DeepSeek-V4.1-Flash~\citep{deepseekv41flash} and GPT-6 Astra~\citep{gpt6} share a turn-based tool-calling interface with identical system instructions and controller. Both models receive the task instruction, robot state information, and head-camera RGB images, and issue tool calls to observe the environment, execute whole-body motions, wait without issuing commands, or finish the episode. Each model waits for the current action to complete before issuing the next call. Both models are accessed through their respective official APIs with medium reasoning effort.

\paragraph{$\Psi_0$.}
We evaluate $\Psi_0$~\citep{wei2026psi_0} using its released checkpoints and native controllers, without fine-tuning on \textsc{LIVIN}.

\subsection{Scene and Object Modelling}

Each scene uses the reconstructed meshes described in Sec.~\ref{sec:construction}, exported from Blender for simulation. Collision geometry is computed with CoACD~\citep{wei2022coacd}, which adaptively decomposes concave meshes into convex hulls. Physical properties (mass, centre of mass, inertia tensor, and contact friction) are initially proposed by GPT-6 Astra, and subsequently reviewed and corrected by the authors.

\subsection{Evaluation Metrics}

Success rate (SR) is the percentage of trials meeting the task-specific success criteria within each task family. For each long-horizon trial, process score (PS) is the percentage of equally weighted physical subgoals completed with their prerequisites satisfied. We report mean PS across trials on a 0--100 scale.